\documentclass[journal,letterpaper]{IEEEtran} 

\usepackage{cite} 
\usepackage{amsmath}
\usepackage{amssymb}
\usepackage{xfrac}
\usepackage{gensymb}

\usepackage{graphicx}
\usepackage[caption=false,font=footnotesize]{subfig} 
\usepackage{tikz}
\usepackage{pgfplots}
\pgfplotsset{compat=1.18}

\usepackage{algorithm}
\usepackage{algpseudocode}

\usepackage{booktabs}
\usepackage{multirow}
\usepackage{array}
\usepackage{longtable}
\usepackage{lscape}

\usepackage[utf8]{inputenc} 
\usepackage[T1]{fontenc}
\usepackage[intoc]{nomencl}
\usepackage{atbegshi}
\usepackage{afterpage}
\usepackage{lineno} 
\usepackage{listings}
\usepackage{enumitem} 
\usepackage{float}

\usepackage[hidelinks]{hyperref}

\usepackage{cite} 
\usepackage{amsmath}
\usepackage{amssymb}
\usepackage{xfrac}
\usepackage{gensymb}

\usepackage{graphicx}
\usepackage[caption=false,font=footnotesize]{subfig} 
\usepackage{tikz}
\usepackage{pgfplots}
\pgfplotsset{compat=1.18}

\usepackage{algorithm}
\usepackage{algpseudocode}

\usepackage{booktabs}
\usepackage{multirow}
\usepackage{array}
\usepackage{longtable}
\usepackage{lscape}

\usepackage[utf8]{inputenc} 
\usepackage[intoc]{nomencl}
\usepackage{atbegshi}
\usepackage{afterpage}
\usepackage{lineno} 
\usepackage{listings}
\usepackage{enumitem} 
\usepackage{float}

\usepackage[hidelinks]{hyperref}

\begin{document}

\title{Fast Fourier Convolutional GAN for 30 m Clear-Sky Land Surface Temperature Gap-Free Reconstruction}

\author{Marwa Alfouly, Smajil Halilovic, Nils Bochow, Niklas Boers, Thomas Hamacher, Konrad Schindler
\thanks{This work was supported by the TUM International Graduate School of Science and Engineering (IGSSE) within the scope of the Innovation Network EarthCare. \textit{(Corresponding author: Marwa Alfouly, e-mail: marwa.alfouly@tum.de.)}}%
\thanks{Marwa Alfouly, Smajil Halilovic, and Thomas Hamacher are with the Chair of Renewable and Sustainable Energy Systems, Technical University of Munich, Germany.}%
\thanks{Nils Bochow is with the Helmholtz Centre for Polar and Marine Research, Alfred Wegener Institute, Germany.}%
\thanks{Niklas Boers is with the Professorship of Earth System Modelling, Technical University of Munich, Germany.}%
\thanks{Konrad Schindler is with the Swiss Federal Institute of Technology Zurich (ETH Zurich), Institute of Geodesy and Photogrammetry, Switzerland.}%
}%

\markboth{Journal of Selected Topics in Applied Earth Observations and Remote Sensing}%
{Shell \MakeLowercase{\textit{et al.}}: A Sample Article Using IEEEtran.bib for IEEE Journals}

\maketitle
\begin{abstract}

Satellite-derived Land Surface Temperature (LST) provides spatially comprehensive data that ground stations cannot match. However, its utility is frequently limited by severe data gaps due to the presence of clouds. As LST is essential for understanding land-atmosphere interactions, numerous methods have been proposed to address this challenge. Yet, the development of a scalable and adaptable pipeline for generating gap-free LST datasets and reconstructing cloud-contaminated pixels remains challenging. Moreover, the reconstruction of extensive missing regions in fine-spatial-resolution observations is particularly difficult. To address this challenge, we propose a Multimodal Fast Fourier Convolutional GAN for reconstructing cloud-contaminated pixels in fine-resolution (30 m) Landsat imagery to generate gap-free clear-sky LST products. The method leverages Fast Fourier Convolution to enable a global receptive field across the image, and is guided by a stack of data consisting of satellite observations and Synthetic Aperture Radar (SAR) data. Across all LST quantiles, the interquartile range of scene-averaged RMSE (computed over reconstructed pixels) is consistently between 0.8 K and 1.8 K. Applied without retraining to an independent region in western Germany, the corresponding interquartile range shifted only moderately to between 1.1 K and 2.2 K, and the median RMSE increased from 1.34 K to 1.58 K. The proposed approach enables the recovery of extensive missing regions, including scenes with more than 70\% cloud-induced gaps, while relying on auxiliary data that are readily available at a near-global scale. 

\end{abstract}


\section{Introduction}\label{sec:Introduction}
\label{sec:Introduction}
\IEEEPARstart{L}{and} Surface Temperature (LST) serves as a foundational thermodynamic parameter governing the energy equilibrium at the surface-atmosphere interface, where it regulates the exchange of long-wave radiation, sensible heat, and latent heat fluxes~\cite{Dash.2002, Thakur.2022}. Beyond its role in atmospheric physics, LST is a widely used indicator across diverse scientific disciplines. For example, it is used in public health to map the habitat suitability of disease vectors like mosquitoes~\cite{Hou.2024, Nayak.2025}, in geology to identify geothermal energy potential~\cite{Putri.2021, Akhyar_Sary_2024}, and in disaster management to detect active wildfires and monitor permafrost degradation~\cite{Batbaatar.2020, Jodhani.2024, Rivera.2026}. In hydrological and agricultural contexts, LST enables the calculation of crop water stress and the detection of invisible early-stage droughts~\cite{ElShirbeny.2017, Ciezkowski.2020}. Monitoring LST is increasingly critical today, as record-breaking near-surface air temperatures in 2024 and 2025 exceeded the 1.5\celsius\ pre-industrial threshold~\cite{C3S.2025, Forster.2026}, reflecting an accelerated warming of the Earth system that is directly manifested in LST patterns. This warming is most acute in metropolitan areas, where rapid urbanization replaces natural vegetation with impervious materials, lowering albedo and increasing thermal inertia~\cite{Vujovic.2021, Yu.2024}. Such transformations, reinforced by the urban canyon effect that traps radiation and restricts convective cooling, result in the Surface Urban Heat Island (SUHI) effect~\cite{Giannopoulou.2010, Khalvandi.2025}. While SUHI mitigation through green and blue infrastructure is vital, its effectiveness is non-linear and scale-dependent. Because the spatial configuration of buildings creates sharp thermal gradients over short distances, monitoring these localized impacts and optimizing urban energy efficiency necessitates high-resolution observational data, typically at a scale of 100 meters or finer, to resolve the thermal signatures of individual city blocks~\cite{Sobrino.2012}.

The acquisition of LST data at regional or global scales relies predominantly on remote sensing observations, using passive microwave (PMW) or thermal infrared (TIR) sensors. TIR sensors measure the spectral radiance emitted by the Earth's surface in the atmospheric window (typically 8--14 $\mu$m), where atmospheric absorption is minimal~\cite{Ghent.2017, Ahmed.2023}. The retrieved at-sensor radiance is subjected to rigorous atmospheric correction and emissivity modeling to invert Planck's law and estimate the true LST~\cite{Ghent.2017}. In contrast, PMW sensors detect naturally occurring microwave radiation at much longer wavelengths (0.8--75 cm)~\cite{Ahmed.2023}. Because the ability of an optical system to resolve detail is limited by diffraction, which is directly tied to wavelength, TIR sensors can achieve relatively high spatial resolutions (30~m--4~km), enabling fine-scale surface temperature analysis. PMW sensors, however, are inherently constrained to much coarser spatial resolutions (typically 10--50 km) because detecting the weaker, longer-wavelength microwave signal requires a large instantaneous field of view. Despite this limitation, microwave observations offer a critical advantage: their ability to penetrate cloud cover. Thermal infrared radiation, in contrast, is strongly affected by clouds. Water droplets and ice crystals absorb the upwelling thermal infrared radiation and re-emit it at lower temperatures corresponding to the cloud-top altitude. Consequently, when a satellite observes a cloud-covered region, the retrieved temperature reflects the cloud top rather than the land surface, rendering the pixel invalid. This challenge is significant, as approximately 67\% of the Earth's surface is obscured by clouds at any given time, creating persistent data gaps, especially in high-humidity and maritime climates~\cite{King.2013}. Beyond the obvious obstruction posed by optically thick clouds, TIR remote sensing is also highly sensitive to sub-pixel cloud contamination, thin cirrus clouds, and cloud shadows~\cite{vermote.1999, Li.2017}. Thin clouds can cause a partial attenuation of the surface thermal signal, leading to subtle but significant underestimations of LST. Furthermore, the shadows cast by clouds significantly lower the surface temperature relative to adjacent sunlit areas, creating transient thermal anomalies that do not reflect the baseline thermodynamic properties of the land cover. 

As a result, significant efforts have been directed towards the generation and validation of gap-free high spatiotemporal resolution LST data~\cite{Li.2013, Li.2021}. These efforts have increasingly focused on the filling and reconstruction of cloud-contaminated images, aiming to recover missing or obscured LST information. Gap-filling methods address either clear-sky LST, which describes the temperature that the surface would have under cloud-free conditions, or all-weather LST, which describes the actual surface temperature under clouds~\cite{Zeng.2018, Zhang.2020, Tan.2021}. Because cloud cover attenuates incoming solar radiation, daytime clear-sky LST exceeds all-weather LST~\cite{Ermida.2019}. Several applications, such as drought monitoring with thermal indices~\cite{Kogan.1995, Sandholt.2002}, the detection of crop water stress~\cite{Idso.1982, Veysi.2017} and the mapping of intra-urban heat exposure~\cite{Hoffman.2020}, are based on clear-sky LST. Other applications, such as evapotranspiration estimation~\cite{Allen.2007, Melton.2022} and annual temperature cycle climatologies~\cite{Bechtel.2015}, can also be based on clear-sky LST, although all-weather LST is preferable when it is available. Clear-sky and all-weather LST are nevertheless complementary rather than competing, since, on the one hand, two-step all-weather frameworks first reconstruct the clear-sky LST and then correct it for the cloud effect~\cite{Zeng.2018, Jia.2021, Zhang.2024a}, and the accuracy of the clear-sky step has been shown to dominate the accuracy of the final cloudy-sky estimate~\cite{Jia.2021}. On the other hand, recent deep-learning methods estimate all-weather LST by adding reanalysis, assimilation or Earth system model fields that contain the cloud effect as additional inputs of the network~\cite{Gong.2023, Zhang.2023, Liu.2025b}, so that a guided clear-sky reconstruction network, such as the one proposed in this study, can be extended to all-weather LST by complementing its guiding stack with such fields and training it against all-sky targets.

All-weather reconstruction has so far been developed mainly for MODIS and geostationary data at 1 km and coarser~\cite{Li.2024a, Jia.2023}. The few methods that extend it to the Landsat scale take the signal under clouds from sources at about 1 km to 30 km, namely all-sky geostationary LST with modeled cloudy pixels~\cite{Wang.2024, Tang.2024}, land surface model simulations~\cite{Ma.2022}, surface energy balance estimates~\cite{Gao.2024} and reanalysis-driven annual temperature cycle models~\cite{Liu.2025a, Liu.2025b}. Under clouds, their spatial detail is therefore limited to the resolution of these sources. Moreover, their accuracy under clouds was assessed at a few in situ stations over rural vegetated or desert surfaces, such as the stations of the US Surface Radiation Budget Network (SURFRAD)~\cite{Zhu.2022, Wang.2024, Tang.2024, Liu.2025b}, cropland, forest and grassland sites in China~\cite{Guo.2025} or low-elevation flux towers in the vegetated part of a mountainous nature reserve~\cite{Zhao.2026}. Urban surfaces, where the LST varies strongly within a few tens of meters and where a large share of the population is exposed to heat, are not represented among these validation sites. Clear-sky reconstruction at the native Landsat resolution thus remains an essential building block, both for clear-sky applications and as the first step toward all-weather LST at 30 m. The present study focuses on the clear-sky target, which can be validated pixel by pixel against true satellite LST observations by applying artificial cloud masks, whereas a reliable extension to all-weather LST at 30 m would require a reference under clouds that goes beyond such sparse point measurements~\cite{Guillevic.2018}.

The development of LST reconstruction methods, ranging from spatial-temporal interpolation to multi-source fusion, has been systematically documented in the literature. Mo et al.~\cite{Mo.2021} and Wu et al.~\cite{Wu.2021} provided comprehensive reviews of progress through 2020, covering established traditional methods and early adoption of machine learning techniques. Building on this, Alfouly et al.~\cite{Alfouly.2026} reviewed advancements through 2025, specifically highlighting a subsequent paradigm shift in which machine learning and deep learning architectures transitioned from emerging tools into the primary, dominant methodology for high-resolution LST data recovery. 

The aforementioned literature reviews strongly support that reconstructing LST under cloudy conditions has become a prominent research topic, with a rapidly growing number of algorithms developed over the past decade. However, the vast majority of this research relies on medium-resolution data, specifically the 1 km daily Terra and Aqua MODIS LST products, which provide highly accessible and frequent observations. In contrast, research directly reconstructing fine-resolution Landsat data (30--100 meters) is much less common because of the inherent trade-off between spatial and temporal resolutions. Since Landsat has an effective revisit cycle of 16 days for a single satellite, or approximately 8 days when combining Landsat-8 and Landsat-9~\cite{USGS_Landsat_C2L2, Li_J.2020}, frequent cloud contamination creates massive temporal gaps that makes it exceptionally difficult to reconstruct missing pixels using only the sensor's own sparse historical data. To overcome this limitation, most fine-resolution research focuses on spatiotemporal fusion techniques, which blend the frequent, coarse-resolution MODIS observations with the detailed but infrequent Landsat observations to generate gap-free, high-resolution daily LST datasets. 

Well-established fusion methods that generate Landsat-like LST include the enhanced adaptive reflectance fusion model (ESTARFM)~\cite{Zhu.2010} and the spatiotemporal adaptive data fusion algorithm for temperature mapping (SADFAT)~\cite{Weng.2014}. However, a key challenge in applying ESTARFM or SADFAT directly to LST fusion is that LST observations from different sensors are often not radiometrically comparable due to differences in retrieval algorithms, viewing geometry, and acquisition times, which introduce systematic biases. More recent fusion models learn these differences from the data, using convolutional and transformer architectures that jointly model sensor differences and temporal change~\cite{Yin.2021, Jin.2026, Wang.2026} or pixel-wise time-series regression~\cite{Chen.2023}. Like ESTARFM and SADFAT, these models are primarily designed for the estimation of clear-sky LST, and since MODIS observes the surface at a similar time of day and is affected by largely the same clouds, they require clear-sky image pairs of both sensors and therefore mainly fill the temporal gaps between Landsat overpasses rather than large overcast patches within a scene~\cite{Yin.2021}.

To avoid the complexities associated with multi-sensor radiometric intercalibration, an alternative class of methods focuses on LST reconstruction using auxiliary guiding data. These approaches leverage variables that exhibit strong physical or statistical correlations with LST, such as Digital Elevation Models (DEMs) for topographic effects, Normalized Difference Vegetation Index (NDVI) for land cover characteristics, or even Synthetic Aperture Radar (SAR) backscatter. For instance, Zhu et al.~\cite{Zhu.2022} proposed a model that utilizes gap-filled Landsat NDVI to guide LST reconstruction. While this approach performs well at the SURFRAD sites, its reliability declines significantly when cloud cover exceeds 70\%. Moreover, the requirement for cloud-free Landsat NDVI introduces an additional constraint that limits its broader applicability. More recent approaches incorporate auxiliary variables that are less sensitive to cloud contamination. For example, Li et al.~\cite{Li.2024} developed a machine learning based framework using eXtreme Gradient Boosting (XGBoost), integrating SAR backscatter and DEM-derived topographic features to guide the LST reconstruction. Because SAR signals can penetrate cloud cover and provide consistent structural information, their SDX-LST model demonstrates more stable performance, even under extreme cloud conditions exceeding 70\%. However, it is strictly designed for vegetated areas. During their experiments, the authors explicitly restricted their regional predictions to trees, grass, crops, and sparse vegetation. All other land cover types (such as urban built-up areas, open water, and barren land) were treated as missing data because the model is not equipped to reconstruct them. Additionally, the authors report that reconstruction accuracy declines with increasing distance between training samples and prediction locations. Their experiments are conducted over relatively localized regions, where spatial gaps remain limited in extent. In contrast, a 70\% gap in a full Landsat scene can span much larger and more heterogeneous areas, requiring predictions over substantially greater distances than those represented in the experimental setup. Under these conditions, the reported performance, derived from localized experiments, may not be representative at broader spatial scales.

Other approaches leverage deep learning architectures, particularly Convolutional Neural Networks (CNNs) and Generative Adversarial Networks (GANs), to specifically reconstruct clear-sky Landsat LST. For instance, Chen et al.~\cite{Chen.2020} introduced the Source Augmented Partial Convolution v2 (SAPC2) model, a CNN-based U-Net architecture designed for Landsat 8 LST inpainting. A key limitation of SAPC2 is its reliance on highly correlated, temporally adjacent LST image pairs, which constrains its applicability under persistent cloud conditions. Rather than training on clear-sky images with synthetically generated masks, SAPC2 depends on a large volume ($\approx$ 2.5 million) of real paired observations, where each target image is matched with a temporally adjacent source image. This design increases data demands and reduces flexibility, particularly in regions with limited clear-sky observations. Another framework by Khedher et al.~\cite{Khedher2.2024} is based on a conditional GAN, which translates non-thermal RGB orthophotography into fine-scale thermal maps, guided by DEM-derived topographic features and land cover data. However, RGB-to-LST approaches inherently require clear-sky RGB inputs, such as pre-captured cloud-free aerial orthophotography, to function. Beyond these data dependencies, a fundamental architectural challenge remains: the restricted receptive fields of standard convolutions often prevent these models from capturing long-range contextual dependencies. 

To address these limitations, this study proposes a new deep learning framework for large-scale Landsat LST reconstruction under extreme cloud contamination. First, we adapt a resolution-robust large-mask inpainting architecture, namely LaMa~\cite{Suvorov.2021}, originally designed for natural RGB images with three-channel inputs, to thermal LST data by incorporating physically meaningful auxiliary guiding variables. LaMa is based on Fast Fourier Convolutions (FFC), which enable image-wide receptive fields and thereby overcome the locality constraints of standard convolutional networks. Second, auxiliary guiding data are systematically integrated and analyzed to assess their contribution to reconstruction performance; importantly, these inputs are derived from publicly available datasets and do not require prior gap-filling. Third, we explicitly target highly challenging scenarios by reconstructing cloud gaps exceeding 70\% of full Landsat scenes (6000 $\times$ 6000 pixels), significantly extending beyond the gap sizes typically considered in prior work. Fourth, model performance is evaluated not only using average error metrics but also across error distributions (quantiles), ensuring that the models capture variability and do not simply regress toward the mean. Fifth, we compare multiple reconstruction strategies, including the SDX-LST framework, the original LaMa model applied to LST data replicated across three channels to match its RGB-based input structure, our adapted LaMa variant tailored for LST reconstruction, and a SwinUNet-based GAN in which the generator in the original LaMa architecture is replaced with SwinUNet~\cite{Cao.2021}. Sixth, we apply the trained models without retraining to an independent Landsat scene footprint in western Germany, which shares the temperate climate of the study area but is on average about 1 K warmer and contains more forest and built-up areas, in order to assess their performance outside the training region. Finally, we release the complete training and inference pipeline as open source at \url{https://github.com/Marwa-AL/Landsat-LST-Reconstruction}, enabling researchers to reproduce our experiments, retrain the framework using publicly available datasets, and adapt it to different geographic regions and remote sensing applications.


This paper is organized as follows. Section~\ref{sec:methodology} describes the study area, data acquisition, and the proposed architecture. Section~\ref{sec:Implementation} presents the implementation details, including the ablation study, testing procedures, and validation protocol. The results are presented and discussed in Sections~\ref{sec:Results} and~\ref{sec:Discussion}, respectively. Finally, Section~\ref{sec:Summary} presents the conclusions.

\section{Methodology}
\label{sec:methodology}
The proposed inpainting framework is systematically developed in three interconnected phases, transitioning from the acquisition of raw satellite signals to a physically consistent thermal reconstruction. Phase I focuses on multi-source data integration and spatial preprocessing, ensuring alignment between thermal, SAR, and vegetation indices (Section~\ref{ssec:Phase1}). Phase II details the network architecture, centered on the use of Fast Fourier Convolutions (FFCs) to capture global thermal structures (Section~\ref{ssec:Phase2}). Finally, Phase III defines the hybrid loss configuration, which combines adversarial learning with physical constraints to preserve urban heat hotspots (Section~\ref{ssec:Phase3}).
\subsection{Study Area and Data Acquisition}
\label{ssec:Phase1}
\subsubsection*{\textbf{Study Area, The Bavarian Macro-Region}}
\begin{figure*}[!t]
\centering
\includegraphics[width=\textwidth]{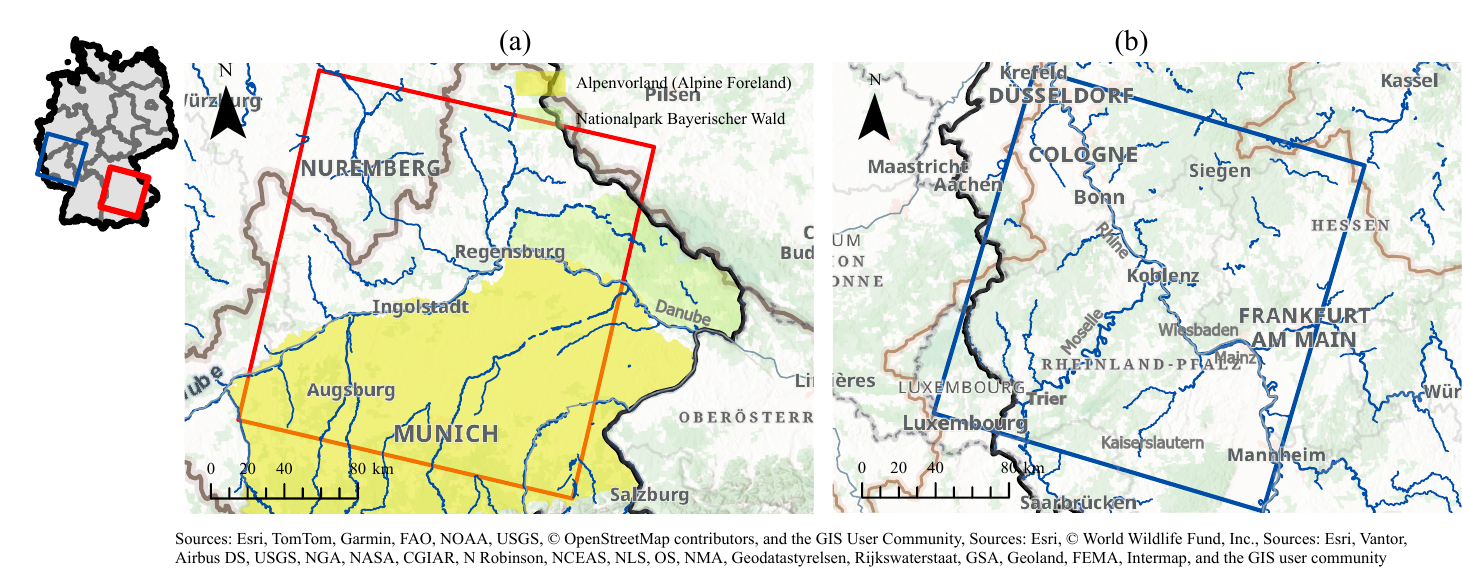}
\caption{Geographical overview of the study area and the independent test region: (a) study area in Bavaria (Landsat WRS-2 path 193, row 26) and (b) independent test region in western Germany (WRS-2 path 196, row 25), shown at the same scale, with the location of both regions in Germany on the left.}
\label{fig:study_area}
\end{figure*}

The study focuses on a representative $185 \times 180 \text{ km}$ region in Bavaria, Germany (UTM Zone 32N), centered at $48^\circ51'$N, $11^\circ35'$E. This area serves as an ideal testbed for multi-modal LST reconstruction due to its extreme heterogeneity in both land cover and topography. Fig.~\ref{fig:study_area}a illustrates the geographical extent of the study region, which features two contrasting environments: dense urban-industrial centers and diverse natural landscapes. The key characteristics of these environments are as follows:

\indent \textbullet\ \textit{Urban-Industrial Hubs:} The region encompasses major European metropolitan centers, including, (i) Munich, a sprawling urban heat island with high-density built-up surfaces; (ii) Nuremberg \& Fürth, a large industrial and commercial conurbation in Middle Franconia; (iii) Ingolstadt \& Regensburg, strategic industrial hubs located along the Danube, characterized by pronounced thermal signatures associated with manufacturing and transportation infrastructure; and (iv) Augsburg, one of Germany's oldest urban centers, providing complex historical urban fabric data. 

\indent \textbullet\ \textit{Diverse Natural Landscapes:} Beyond the urban centers, the study area encompasses a diverse mosaic of land-cover and physiographic features that strongly influence LST, including, (i) the Danube River Valley, which acts as a major hydrological and thermal regulator extending across the region from west to east; (ii) the Bavarian Forest and Alpine Foreland, comprising extensive spruce- and beech-dominated forests that provide a relatively stable thermal baseline compared with the highly variable temperatures of surrounding agricultural landscapes; (iii) the agricultural heartlands, including the Gäuboden and Hallertau regions, where intensive croplands and meadows (notably hop cultivation) exhibit pronounced seasonal variability in LST; and (iv) the topographic gradient, where the terrain transitions from the flat Danube plains to the rolling Franconian Alb and the Alpine foothills, motivating the inclusion of hillshade and digital elevation model (DEM) layers (Stage 3) to account for topographic shading effects on surface temperature.

\subsubsection*{\textbf{Independent Test Region}}
To assess the performance of the trained models outside the training region, a second region in western Germany, covered by WRS-2 path 196, row 25, is used exclusively for testing (Fig.~\ref{fig:study_area}b). It does not overlap with the Bavarian study area and follows the Rhine from the Cologne--Bonn and Düsseldorf area through Koblenz to the Rhine--Main area with Mainz and Wiesbaden and to Mannheim--Ludwigshafen, so that it combines large urban agglomerations, river corridors, croplands and forested low mountain ranges. Since none of its scenes are used for training and its climate is comparable to that of the study area, it shows how the trained models perform in an unseen region under similar climatic conditions. Nevertheless, the independent region is milder, since the federal states it covers had mean annual air temperatures of 9.3~$^\circ$C to 10.0~$^\circ$C in 1991--2020, compared with 8.6~$^\circ$C for Bavaria~\cite{DWD.2026}, and it contains more forest and built-up areas and less cropland, which account for 41\%, 15\% and 36\% of the reconstructed pixels, compared with 36\%, 10\% and 51\% in the study area. Since the framework is not intended to be applied directly to arbitrary regions, we recommend retraining or fine-tuning the model with local data for regions with substantially different climates, such as arid, tropical or snow-dominated areas, where the relationships between LST and the auxiliary variables may differ. Such retraining is facilitated by the global availability of all auxiliary inputs and by the openly available code for data preparation and training.

\subsubsection*{\textbf{Data Acquisition}}
\label{sssec:data_acquisition}
The design is focused on generating gap-free, Landsat-like LST using a multi-channel composite input, where the primary LST observation is augmented by a stack of auxiliary data for guidance. The data pipeline is designed to provide the model with a comprehensive physical context using readily available open-source, cloud-free, and atmospherically corrected products that require only geographical alignment and normalization. All multi-source auxiliary layers were systematically resampled and geolocated to match the 30 m Landsat grid, ensuring pixel-level alignment across the input stack. Table~\ref{tab:aux_data} lists all inputs with their number of input channels, spatial resolution, preprocessing steps and acquisition and access. To optimize the feature space, our training data preparation followed an additive hierarchical structure across three stages: Stage 1 established the baseline with LST and land use; Stage 2 integrated topographic (DEM) and vegetation (NDVI) indices; and Stage 3 finalized the feature set by incorporating hillshade and dual-polarization radar (SAR) data. 

\indent \textbullet\ \textit{Stage 1: Baseline Inputs} \\The primary dataset comprises Landsat 8/9 Collection 2 Tier 1 Level-2 surface temperature observations and their corresponding cloud masks, specifically targeting Path 193, Row 26 of the Worldwide Reference System-2 (WRS-2) from January 2013 to May 2025~\cite{USGS_Landsat_C2L2}. This data provides a 30 m spatial resolution and a combined 8-day temporal revisit frequency (utilizing both Landsat 8 and 9), which is critical for monitoring urban thermal dynamics. To ensure a high-fidelity training environment, the data underwent a two-step quality control process. First, at the scene level, we retained any Landsat 8/9 Tier 1 overpass with at least 20\% valid pixels (maximum 80\% cloud cover), maximizing the temporal breadth of the archive to 196 scenes. The dataset was then systematically processed using a dual-stage tiling strategy to maximize spatial coverage and feature diversity, beginning with a sliding window of $256 \times 256$ pixels with a 128-pixel stride to ensure 50\% spatial overlap for continuity. To further enhance the robustness of the training set and reduce spatial bias, this systematic approach was supplemented by stochastic sampling of random tiles across the diverse landscape. Second, at the tile level, both the sliding window and random sampling were strictly constrained by the Landsat QA\_PIXEL band; only tiles that were 100\% clear-sky and void of no-data artifacts were admitted into the final training pool, resulting in approximately 36,400 ground-truth tiles for the inpainting framework.

To provide the model with sharp structural guidance, we integrated Esri (Sentinel-2) Land Use/Land Cover (LULC) data at 10 m resolution~\cite{Karra.2021}, featuring 9 distinct classes. For effective normalization and feature representation, the LULC data was one-hot encoded into 9 distinct binary channels, representing the primary land-cover classes of the study area. Furthermore, to capture the cyclical phenological and solar cycles, the Day of Year (DOY) was normalized using sine and cosine transformations, providing the model with a continuous seasonal coordinate.

\begin{table*}[!t]
\centering
\footnotesize
\caption{Input data of the reconstruction framework with the number of input channels, spatial resolution, preprocessing steps and acquisition and access. All inputs are publicly available and are only resampled to the 30 m Landsat grid, not gap-filled or downscaled.}
\label{tab:aux_data}
\renewcommand{\arraystretch}{1.15}
\begin{tabular}{>{\raggedright\arraybackslash}p{2.6cm}>{\raggedright\arraybackslash}p{1.9cm}>{\raggedright\arraybackslash}p{6.6cm}>{\raggedright\arraybackslash}p{4.9cm}}
\toprule
Data (input channels) & Spatial resolution & Preprocessing steps & Acquisition and access \\
\midrule
LST (1) & \multirow[t]{5}{=}{100~m (TIRS), distributed at 30~m} & Conversion to K ($\times 0.00341802 + 149.0$), missing pixels set to zero, normalization to 260--342 K & \multirow[t]{5}{=}{Landsat 8/9 Collection 2 Tier 1 Level-2 surface temperature (ST\_B10) and QA\_PIXEL, USGS~\cite{USGS_Landsat_C2L2}} \\
Binary mask (1) & & Pixels flagged as cloud, cloud shadow, dilated cloud, cirrus or snow in QA\_PIXEL and pixels without data, combined with the artificial cloud masks & \\
\addlinespace[4pt]
Land cover (9) & 10~m, annual & Map of the acquisition year or, if not available, of the closest available year, nearest-neighbor resampling, one-hot encoding of 9 classes & Esri annual 10 m Sentinel-2 LULC~\cite{Karra.2021}, Esri Land Cover Explorer \\
\addlinespace[4pt]
NDVI (1) & 300~m, 10-daily composites & Composite closest to the acquisition date in the acquisition year, or at the same time of year in the closest available year if the acquisition year is not covered (2013), bilinear resampling, scaling ($\times 0.004 - 0.08$), normalization from $-1$ to 1 & CLMS NDVI 300 m, V1 and V2~\cite{CLMS_NDVI_300m_V1, CLMS_NDVI_300m_V2}, Copernicus Land Monitoring Service \\
\addlinespace[4pt]
Elevation (1) & \multirow[t]{3}{=}{30~m, static for all scenes} & Reprojection to UTM zone 32N on the Landsat grid (bilinear), normalization from $-500$ to 9000~m & \multirow[t]{3}{=}{Copernicus GLO-30 DEM~\cite{ESA.2024} with slope and aspect, distributed by OpenTopography} \\
Slope (1) & & Reprojection to UTM zone 32N on the Landsat grid (bilinear), normalization from 0 to 90$^\circ$ & \\
Aspect (2) & & Reprojection to UTM zone 32N on the Landsat grid (nearest neighbor), sine and cosine encoding & \\
\addlinespace[4pt]
Hillshade (1) & 30~m & Computed for each overpass from the slope and aspect of the DEM and the sun elevation and azimuth angles with the cosine law of illumination (Eq.~\ref{eq:hillshade}), reprojection to the Landsat grid (nearest neighbor), normalization from 0 to 255 & Derived from the GLO-30 DEM and the sun angles in the MTL metadata file downloaded with each Landsat scene, USGS~\cite{USGS_Landsat_C2L2} \\
\addlinespace[4pt]
Sentinel-1 VV (1) & \multirow[t]{2}{=}{10~m, exported at 30~m} & Median of all acquisitions within $\pm 7.5$ days of the Landsat overpass ($\pm 25$ days if the footprint is not fully covered), incidence angles of 31$^\circ$--45$^\circ$, export on the Landsat grid, normalization to $-25$ to 5 dB, filled with $-25$ dB for acquisitions before the start of Sentinel-1 & \multirow[t]{2}{=}{Sentinel-1 GRD, IW mode, descending orbit, Google Earth Engine~\cite{ESA_Sentinel1_GEE, Gorelick_GEE_2017}} \\
Sentinel-1 VH (1) & & As for VV, normalization to $-30$ to $-5$ dB, filled with $-30$ dB before the start of Sentinel-1 & \\
\addlinespace[4pt]
Day of year (2) & -- & Sine and cosine encoding with a period of 366 days & Landsat acquisition date \\
\bottomrule
\addlinespace[3pt]
\multicolumn{4}{>{\raggedright\arraybackslash}p{17.2cm}}{\footnotesize Note: The Python script for the export of the Sentinel-1 composites on the Landsat grid from Google Earth Engine and the scripts for the preprocessing of all inputs, the computation of the hillshade, the tiling and the generation of the training data are provided in the code repository (see Code Availability), so that only the raw data listed in the table have to be obtained.} \\
\end{tabular}
\end{table*}

\textbullet\ \textit{Stage 2: Topographic and Vegetation Indicators} \\To characterize surface vegetation, we integrated 10-daily composite NDVI data from the Copernicus Land Monitoring Service at a 300 m resolution~\cite{CLMS_NDVI_300m_V1, CLMS_NDVI_300m_V2}. While this product provides 10-day near-real-time estimates globally from January 2014 onwards, our Landsat archive extends back to 2013. To maintain a complete input stack for the entire study period, we implemented a temporal proxy strategy for missing dates. In cases where concurrent NDVI was unavailable, specifically for 2013 observations, we utilized the chronologically closest available DOY from the subsequent years (e.g., substituting a missing DOY 70 in 2013 with the nearest available day in 2014). Although simple resampling of 300 m NDVI might limit the capture of fine-scale spatial variations, it serves as a robust baseline to mitigate intra-class variability within vegetated areas. 
    
\textbullet\ \textit{Stage 3: Multi-Modal Fusion (SAR and Terrain)} \\The auxiliary data stack is further advanced with a dynamic hillshade layer ($H$). Unlike the static terrain layers, this illumination index was independently computed for each Landsat overpass using the following cosine law of illumination~\cite{Burrough.1998, esri_hillshade_2021}:
    \begin{equation} \label{eq:hillshade}
    \begin{split}
    H &= 255 \Big( \cos(Z)\cos(S) \\ 
    &\quad +\sin(Z)\sin(S)\cos(A_z - A_s) \Big) 
    \end{split}
    \end{equation}
where $Z$ is the Sun Zenith angle (derived as $90^\circ - \text{Solar Elevation}$ from metadata), $A_z$ is the Sun Azimuth, $S$ is the terrain slope, and $A_s$ is the terrain aspect. This multi-layered topographic approach allows the model to differentiate between constant altitude-dependent thermal gradients and transient, solar-induced shading patterns. To account for surface roughness and moisture dynamics, we integrated Sentinel-1 Synthetic Aperture Radar (SAR) observations~\cite{ESA_Sentinel1_GEE}. We specifically utilized the descending pass (captured at 06:00 local time), as its temporal proximity to the Landsat overpass (10:00 local time) provides a more accurate representation of morning soil moisture and vegetation structure compared to the ascending evening pass. We retrieved the Level-1 Ground Range Detected (GRD) product directly from Google Earth Engine (GEE)~\cite{Gorelick_GEE_2017}, which provides data that has already undergone standardized radiometric and geometric corrections. Specifically, the dual-polarization VV and VH bands were provided as Analysis-Ready Data (ARD) having already been processed for Radiometric Calibration (converting Digital Numbers DNs to backscatter $\sigma^0$) and Thermal Noise Removal. Crucially, the GEE pipeline includes Range-Doppler Terrain Correction; this step utilizes a Digital Elevation Model to correct the radar imagery, ensuring that every SAR pixel is inherently aligned with the 30 m Landsat LST grid.

While the GEE pipeline includes Range-Doppler Terrain Correction to align the SAR pixels with the 30 m Landsat LST grid, we acknowledge that the extreme topography in the southernmost Alpine fringe of our study area introduces inherent radiometric distortions. Specifically, slopes oriented toward the radar sensor exhibit artificially high backscatter, while those facing away appear disproportionately dark. By providing the model with raw elevation, slope, and aspect alongside these SAR channels, we enable the neural network to learn to compensate for these topographic-radiometric biases, effectively decoupling terrain-induced backscatter fluctuations from the physical moisture and vegetation signals.

\begin{figure*}[ht!]
\centering
\vspace*{0.1cm} 
\scriptsize
\centering
\includegraphics[width =\textwidth]{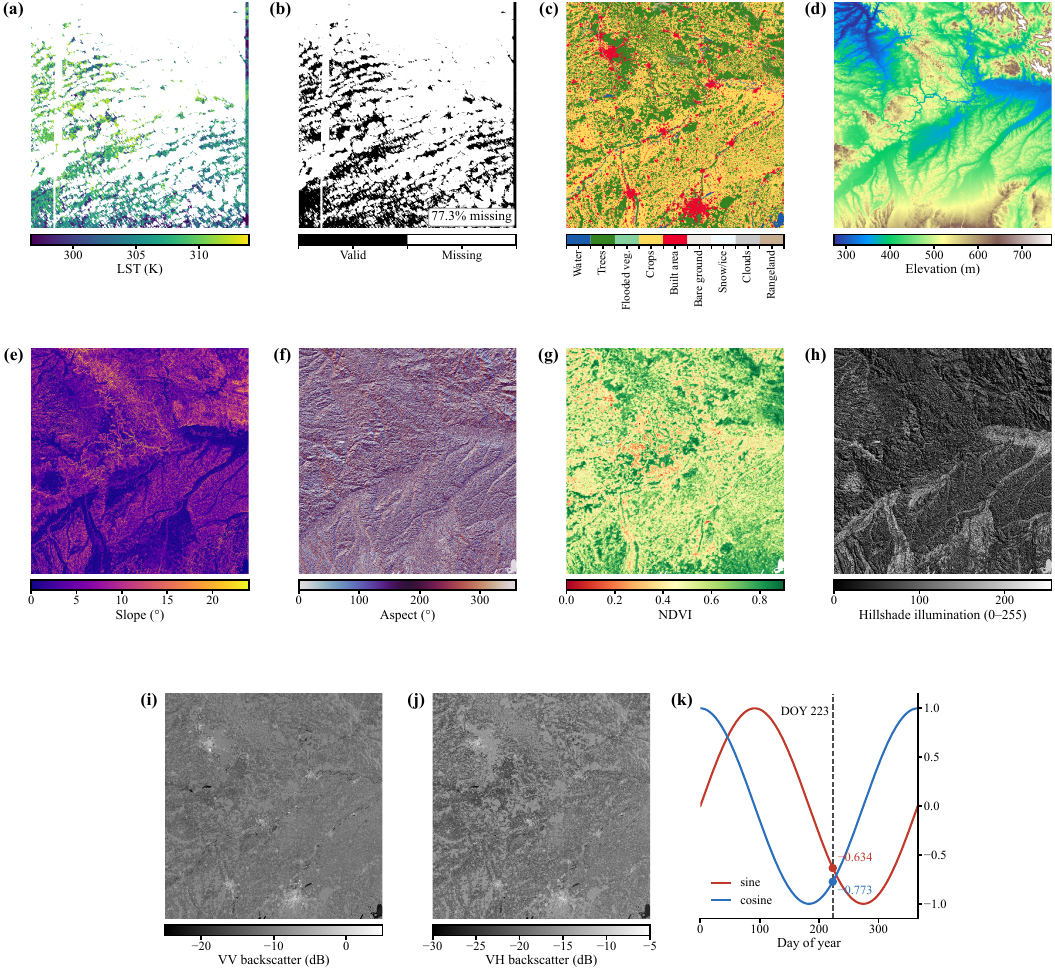}
\caption{\footnotesize Unnormalized input channels for the scene of August 11, 2022: (a) Land Surface Temperature (LST) with the missing pixels shown in white, (b) binary mask of the missing pixels, (c) land use/land cover (LULC) classes, (d) elevation, (e) slope, (f) aspect, (g) Normalized Difference Vegetation Index (NDVI), (h) hillshade, (i) Sentinel-1 VV and (j) Sentinel-1 VH backscatter, and (k) sine and cosine encoding of the day of year (DOY), which is constant within a scene and is therefore shown over the year with the date of the scene (DOY 223) marked.}
\label{fig:aux_data}
\end{figure*}

Fig.~\ref{fig:aux_data} shows the complete input stack for the scene of August 11, 2022. The aspect is shown in degrees, whereas the model receives its sine and cosine, and the DOY encoding, which has the same value for all pixels of a scene, is shown as its annual cycle with the date of the scene marked.

\subsection{Network Architecture: Multi-Modal FFC GAN}
\label{ssec:Phase2}
\begin{figure*}[h]
\centering
\scriptsize
\centering
\includegraphics[width =\textwidth]{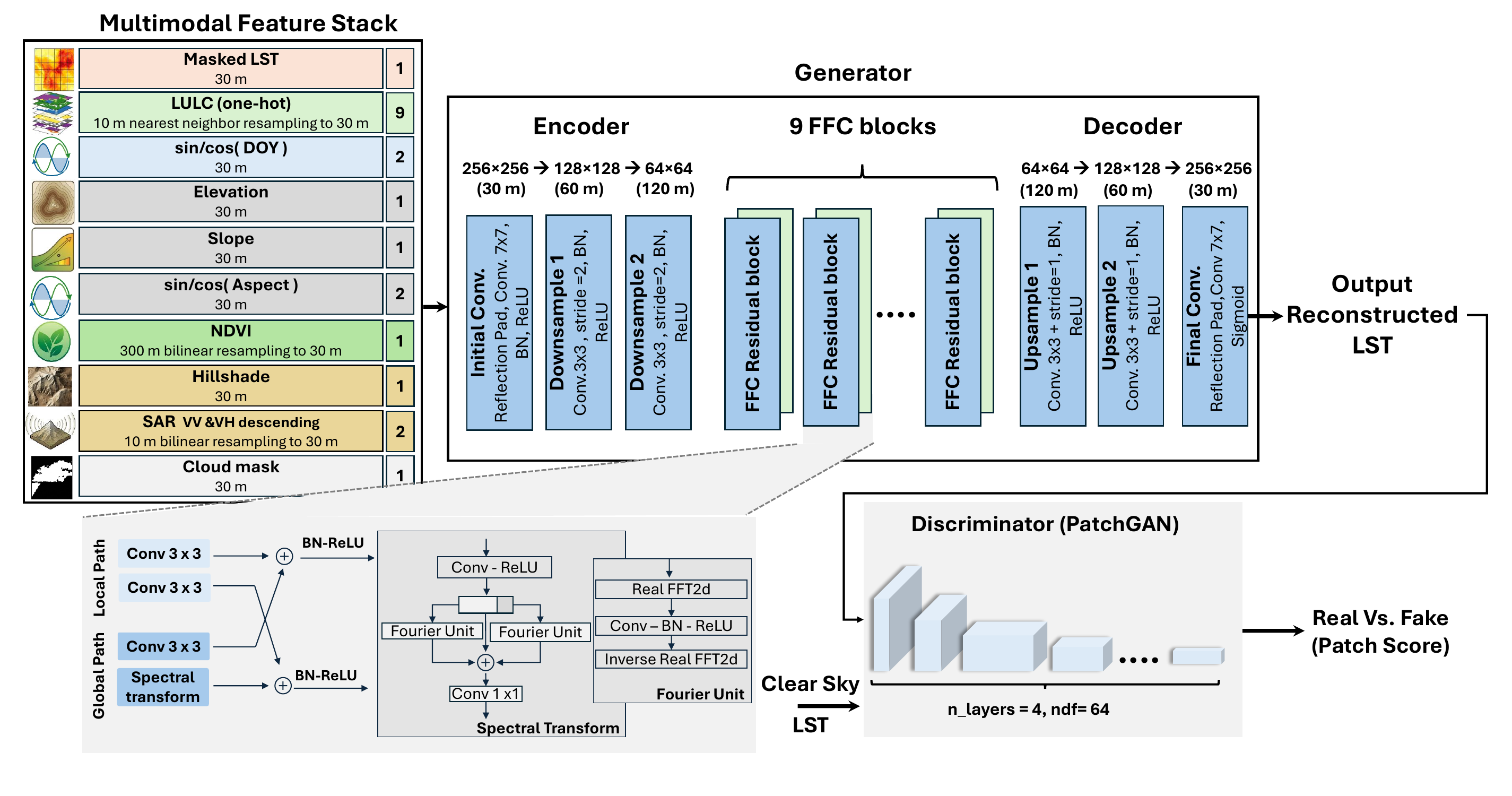}
\caption{The overall architecture of the LST reconstruction model.}
\label{fig:schematic}
\end{figure*}

The proposed model utilizes a Generative Adversarial Network (GAN) framework optimized for large-scale spatial inpainting of LST. This network is an adaptation of the LaMa model~\cite{Suvorov.2021}, which was originally designed for resolution-robust inpainting in RGB images and has shown success in coarse-resolution climate data~\cite{Bochow.2025}. We extend this framework to handle the complex physical variables of Earth observation, transitioning from a simple RGB space to a high-dimensional feature environment. As illustrated in Fig.~\ref{fig:schematic}, the system consists of a generator ($G$) designed for global-local feature synthesis and a discriminator ($D$) that enforces spatial and radiometric realism. 

\subsubsection*{\textbf{Generator Architecture}}
\label{sssec:generator_arch}
To reconstruct missing LST values, the generator follows a modified U-Net backbone, integrated with FFC. While standard convolutional layers are constrained by a local receptive field, the FFC layers utilize a spectral transform to capture global context and long-range spatial dependencies. This is particularly critical for LST reconstruction, as it allows the model to leverage distal topographic and land-cover correlations to predict surface temperatures even within extensive cloud-contaminated regions. By transitioning from the standard three-channel RGB space to this high-dimensional feature environment, the model learns to synthesize radiometrically consistent $30\text{ m}$ thermal signals from diverse environmental drivers. As illustrated in Fig.~\ref{fig:schematic}, the architecture processes the complex input through a symmetrical encoder-decoder structure with a deep spectral bottleneck, consisting of 15 functional blocks organized into five distinct stages. To optimize the feature space, the model is designed to integrate an additive hierarchical feature set based on the data availability stages defined in Section~\ref{sssec:data_acquisition}: 

\begin{itemize}[leftmargin=3em, itemsep=3pt, topsep=6pt]
    \item Baseline Configuration: Processes LST observations, one-hot encoded LULC (9 channels), and cyclical temporal encodings (sin/cos DOY).
    \item Topographic-Vegetation Expansion: Incorporates NDVI and primary terrain attributes (Elevation, Slope, and sin/cos Aspect).
    \item Full Multi-Modal Fusion: Finalizes the stack by incorporating dynamic hillshade and dual-polarization SAR (VV/VH) channels. 
\end{itemize}

The encoder comprises Stages 1--2 and serves as the entry point of the generator. Stage 1 consists of an initial convolution layer with a $7 \times 7$ kernel and reflection padding, projecting the multimodal input stack into a 64-dimensional latent space. This relatively large kernel is employed to increase the initial receptive field and facilitate the aggregation of a broader local spatial context between LST and the auxiliary drivers. Subsequently, Stage 2 performs downsampling through two convolutional blocks, each using a $3 \times 3$ kernel with stride 2, followed by Batch Normalization (BN) and ReLU activation, which reduces the size of the feature maps by a factor of four in each direction, so that the effective ground sampling distance of the features increases from $30\text{ m}$ to $120\text{ m}$.

The core of the architecture (Stage 3) consists of 9 sequential FFC Residual Blocks operating at the bottleneck resolution. Unlike standard convolutions that are limited by local receptive fields, the FFC blocks split the signal into a spatial path for local features and a spectral path for global features. The spectral path utilizes a Fast Fourier Transform (FFT) to perform convolutions in the frequency domain, allowing the model to exploit non-local clear-sky information from across the entire $256 \times 256$ scene to inpaint cloudy regions. Residual connections are employed across each block to ensure gradient stability, enabling the model to learn the complex, long-range correlations between distal topography and the masked LST pixels.

The decoder comprises Stages 4 and 5, which progressively restore spatial resolution and generate the final reconstructed LST map. Stage 4 utilizes a Resize-Convolution strategy consisting of two sequential blocks of bilinear upsampling (factor of 2) followed by a $3 \times 3$ convolution with a stride of 1. This design was selected over transposed convolutions to prevent checkerboard artifacts and ensure the spatial continuity of reconstructed thermal gradients across cloud-covered regions. Consistent with the encoder design, the decoder blocks employ ReLU activations and BN, with BN stabilizing adversarial training and facilitating convergence. The decoder then concludes with Stage 5, which performs the final image synthesis using a Projective Convolution layer. This terminal stage utilizes a $7 \times 7$ kernel with reflection padding to map the 64-dimensional latent features back to the physical domain, producing a 1-channel output. A Sigmoid activation function is applied to the final feature map to bound the output, which is subsequently unnormalized to obtain the reconstructed LST map at the original $30\text{ m}$ resolution. This final transformation preserves the structural sharpness of the urban fabric and the complex physical temperature gradients driven by underlying topography and surface moisture conditions in the gap-filled product.

\subsubsection*{\textbf{Discriminator Architecture}}
\label{sssec:ddiscriminator_arch}
To ensure the generated LST maps are consistent with ground-truth observations, we implement an N-Layer Discriminator based on the PatchGAN~\cite{Isola.2017} architecture. Unlike standard GAN discriminators that condense an entire image into a single scalar, the PatchGAN classifies individual local patches as real or fake. This localized feedback is essential for LST reconstruction, as it forces the generator to maintain high-frequency structural details and realistic local thermal variances that align with the underlying land-cover and topographic drivers. In our implementation, we utilize the configuration of LaMa~\cite{Suvorov.2021} with $N=4$ and a base filter count $\mathrm{ndf}$ of 64. The architecture consists of a series of six $4 \times 4$ convolutional layers, four with a stride of 2, progressively increasing the feature depth from 64 to 512 channels, and two with a stride of 1, the last of which produces the single-channel output map. Each layer except the first and the last is followed by BN, and all except the last by LeakyReLU activations. In contrast to the original LaMa, where the discriminator judges three-channel RGB images and is updated at every iteration, our discriminator judges only the single-channel LST map, while the generator has to learn the complex mapping between the multimodal auxiliary stack and the thermal output. Since such a discriminator can become an expert too quickly, providing vanishing gradients that prevent the generator from learning, it is updated only at every second training iteration and with a lower learning rate than the generator, which achieves a more balanced adversarial competition.

\subsection{Hybrid loss configuration}
\label{ssec:Phase3}
The training of the Multimodal FFC-GAN is guided by a composite objective function designed to ensure both global radiometric accuracy and local structural realism. To achieve a balance between pixel-wise fidelity and adversarial sharpness, the total objective is defined as:
\begin{equation}
\begin{split}
\mathcal{L}_{final} &= \kappa \mathcal{L}_{Adv}^{G} + \alpha \mathcal{L}_{HRFPL} + \beta \mathcal{L}_{DiscPL} + \lambda L_1 \\ 
&\quad + \delta \mathcal{L}_{gradient} + \gamma R_1
\end{split}
\end{equation}
where the individual components and their respective weights are defined as follows:
\begin{itemize}[leftmargin=3em]
      \item $\kappa \mathcal{L}_{Adv}^{G}$ ($\kappa = 10$): The GAN Adversarial Loss, which encourages the generator to produce naturally looking local details.
      \item $\alpha \mathcal{L}_{HRFPL}$ ($\alpha = 30$): The High-Resolution Perceptual Loss, utilizing a pre-trained VGG-19 backbone~\cite{Simonyan.2015}. Rather than penalizing pixel-level errors, this term matches deep multi-scale feature representations to ensure complex spatial patterns and semantic thermal structures are accurately reconstructed.
      \item $\beta \mathcal{L}_{DiscPL}$ ($\beta = 100$): The Discriminator-based Feature Matching Loss, which stabilizes training by matching intermediate structural statistics between the generated and target distributions.
      \item $\lambda L_1$ ($\lambda = 1.0$): The Pixel-wise Absolute Difference, ensuring strict radiometric reconstruction and absolute scale preservation.
      \item $\delta \mathcal{L}_{gradient}$ ($\delta = 1.0$): The LST Gradient Matching Loss, which enforces structural and textural alignment between the predicted and target LST gradients.
      \item $\gamma R_1$ ($\gamma = 0.001$): The Gradient Penalty, a regularization term applied to the discriminator to penalize sharp changes in gradients and prevent training instability. 
\end{itemize}
The following subsections explain the technical rationale behind these components and the domain-specific adaptations required for the inpainting of complex remote sensing data such as LST.

\subsubsection*{\textbf{Adversarial Learning and Training Stability}}

The adversarial component, $\mathcal{L}_{Adv}$, establishes a competitive optimization between the FFC-based generator ($G$) and the PatchGAN discriminator ($D$). The objective is to encourage the generator to reconstruct cloud-obscured regions while producing LST maps that are locally consistent with the spatial characteristics of the ground-truth observations. This interaction is formulated as the following zero-sum min-max game:

\begin{equation}
\begin{split}
\min_{G} \max_{D} \mathcal{L}_{Adv}(G, D)& = \mathbb{E}_{LST_{gt}} [\log D(LST_{gt})] \\ &\quad + \mathbb{E}_{x_{in}} [\log(1 - D(G(x_{in})))]
\end{split}
\end{equation}
where $LST_{gt}$ denotes the ground-truth LST map, $x_{in}\in\mathbb{R}^{H\times W\times C}$ represents the multimodal input stack provided to the generator, and $G(x_{in})$ denotes the generator prediction for the missing regions. The input stack, $x_{in}$, contains the masked LST observations and several auxiliary variables, where the LST channel is explicitly masked to simulate cloud-induced data loss. The binary mask, denoted by $M$, identifies pixels requiring reconstruction ($M=1$), while valid observations are preserved ($M=0$). Unlike the LST channel, the auxiliary inputs (LULC, NDVI, terrain variables, and SAR) remain unmasked, providing complete structural and environmental guidance throughout the reconstruction process. The input depth (C) varies according to the experimental configuration:

\begin{itemize}[leftmargin=3em, itemsep=3pt, topsep=6pt]
    \item Baseline Configuration ($C=13$): Includes masked LST, binary mask, 9-class LULC, and DOY.
     \item Topographic-Vegetation Expansion ($C=18$): Adds NDVI and primary terrain attributes (Elevation, Slope, Aspect).
     \item Full Multi-Modal Fusion ($C=21$): Finalizes the stack with hillshade and dual-polarization SAR data.
\end{itemize}
In practice, the adversarial optimization is implemented using the non-saturating GAN formulation, where the discriminator and generator are updated alternately. The discriminator is optimized to distinguish ground-truth LST maps from reconstructed LST maps according to

\begin{equation}
\mathcal{L}_{D}
=
-\mathbb{E}[\log D(LST_{gt})]
-
\mathbb{E}[\log(1-D(\hat{LST}))]
+
\gamma R_{1},
\end{equation}
The corresponding adversarial contribution to the generator objective is:

\begin{equation}
\mathcal{L}_{Adv}^{G}
=
-\mathbb{E}[\log D(\hat{LST})].
\end{equation}
Here, $\hat{LST}$ denotes the reconstructed (inpainted) LST map, obtained by combining the generator prediction within the masked regions with the valid observations outside the mask,

\begin{equation}
\hat{LST}
=
LST_{gt}\odot(1-M)
+
G(x_{in})\odot M.
\end{equation}
To stabilize adversarial training, the discriminator is regularized using the $R_{1}$ gradient penalty,

\begin{equation}
R_{1}
=
\mathbb{E}_{LST_{gt}}
\left[
\left\|
\nabla_{LST_{gt}}D(LST_{gt})
\right\|_{2}^{2}
\right],
\end{equation}
which penalizes large discriminator gradients on real samples, resulting in smoother optimization and more stable convergence.

This adversarial loss is incorporated into the overall generator objective using the weighting coefficient $\kappa$, while the discriminator regularization is controlled by $\gamma$. Based on empirical sensitivity analysis, $\kappa=10$ was selected as it provided the optimal balance between enforcing local spatial realism and preserving the radiometric consistency of the reconstructed LST fields. Increasing the adversarial weight (e.g., $\kappa=15$) resulted in excessive discriminator influence, leading to reduced reconstruction accuracy and deviations from the expected thermal patterns. The $R_1$ gradient penalty coefficient was fixed at $\gamma=0.001$, providing stable discriminator feedback throughout training without limiting the generator's ability to learn fine-scale thermal structures.

\subsubsection*{\textbf{Radiometric and Gradient Guidance for Remote Sensing}}
A major difference from standard computer vision inpainting is our heavy reliance on explicit radiometric constraints to satisfy physical laws. In standard image completion, such as the original LaMa task, visual plausibility is the primary goal, and conventional supervised losses are often minimized to avoid blurry results. To achieve accurate LST reconstruction, our approach introduces two additional loss components that enforce radiometric accuracy and spatial consistency:
\begin{itemize}[leftmargin=3em, itemsep=3pt, topsep=6pt]
    \item Integration of $L_1$ Loss: Unlike the original LaMa model which prioritizes perceptual and adversarial terms to avoid blurring, we explicitly integrate a Pixel-wise $L_1$ Loss ($\lambda=1.0$). This adaptation is important for remote sensing; by enforcing a 1:1 mapping in the normalized feature space, we ensure the model preserves the radiometric integrity across the full $260\text{--}342\text{ K}$ physical range, preventing the value drift common in purely adversarial optimizations.
    \item LST Gradient Matching Loss ($\mathcal{L}_{gradient}$): We introduce a specialized gradient component ($\delta=1.0$) that targets the spatial derivatives of the LST signal itself. This ensures that the reconstructed thermal transitions accurately mirror the sharp edges found at urban-rural interfaces, providing a layer of structural enforcement that complements the generator's internal spectral reasoning.
\end{itemize}
\subsubsection*{\textbf{Perceptual and Feature Matching}}
\label{sssec:windows}
The final two complementary loss components are a high receptive field perceptual loss ($\mathcal{L}_{HRFPL}$) and a discriminator-based feature matching loss ($\mathcal{L}_{DiscPL}$).
$\mathcal{L}_{HRFPL}$ is introduced to enforce high-level perceptual consistency by leveraging feature representations extracted from a pretrained network with a large receptive field. Unlike pixel-wise losses, this loss focuses on capturing global spatial dependencies and contextual patterns within the reconstructed LST fields. In this work, a VGG-19 backbone~\cite{Simonyan.2015} is adopted as the feature extractor. Although the original LaMa framework employs a ResNet-50 backbone for its perceptual loss component~\cite{Suvorov.2021}, our experimental evaluation demonstrates that VGG-19 provides better performance for LST reconstruction. This observation is also consistent with previous studies in remote sensing image reconstruction and super-resolution, where VGG-19 has been used as a perceptual feature extractor due to its ability to preserve fine spatial textures and local structural variations~\cite{Bello.2020,Chung.2023}. The selection of VGG-19 is motivated by the characteristics of Earth observation imagery, where thermal information is represented through continuous spatial variations and highly textured patterns, including complex urban environments, agricultural structures, and localized thermal anomalies. VGG-19 feature representations, which are inherently texture-biased, are therefore better aligned with the characteristics of LST fields than networks primarily optimized for high-level semantic abstraction. In contrast, the ResNet-50 backbone used in the original LaMa architecture is designed to extract hierarchical semantic representations and is commonly used in recognition and segmentation tasks. While such semantic feature representations are beneficial for conventional image completion problems involving distinct objects, they are less suitable for continuous physical variables such as LST, where information is primarily encoded through gradual pixel-to-pixel thermal transitions rather than discrete object structures. The weighting coefficient $\alpha$ was finalized at 30 after sensitivity testing across $\alpha \in \{1, 10, 30, 40\}$, where lower values failed to provide sufficient structural contextual flow and $\alpha=40$ proved counterproductive by prioritizing abstract feature patterns over absolute radiometric accuracy.

In addition to $\mathcal{L}_{HRFPL}$, $\mathcal{L}_{DiscPL}$ is introduced to further enhance reconstruction accuracy, supporting structural alignment with the spatial patterns underlying urban and environmental thermal distributions. Unlike conventional perceptual losses that rely on fixed pretrained networks, $\mathcal{L}_{DiscPL}$ utilizes intermediate feature representations extracted from the discriminator during adversarial training. It is defined as:
\begin{equation}
\mathcal{L}_{DiscPL} = \mathbb{E} \sum_{i=1}^{K} \frac{1}{M_i} ||D^{(i)}(LST_{gt}) - D^{(i)}(\hat{LST})||_1
\end{equation}

Here, $K=5$ denotes the number of hidden discriminator layers used for feature matching, and $M_i$ represents the number of elements in the feature representation extracted from the $i$-th discriminator layer.

This loss encourages the generator to align intermediate discriminator feature representations extracted from the ground-truth and reconstructed LST fields, thereby improving structural realism beyond pixel-level supervision. A weighting coefficient of $\beta = 100$ was selected after comparative evaluation against lower values of 30 and 60, which were insufficient to stabilize large-scale structural reconstruction.

Overall, the combination of $\mathcal{L}_{HRFPL}$ and $\mathcal{L}_{DiscPL}$ improves reconstruction quality by incorporating both pretrained and discriminator-derived feature representations.

\section{Experiments}\label{sec:Implementation}
This section details the experimental framework used to evaluate the Multimodal FFC-GAN for the LST reconstruction. The experiments are structured to assess the influence of multi-source data fusion and loss function configurations on radiometric and structural accuracy.
\subsection{Experimental Variants}\label{ssec:variants}

The core of the experimental framework is a systematic ablation study designed to evaluate the FFC-GAN’s behavior across varying levels of environmental guidance and loss configurations. This phased assessment allows for a progressive analysis of how the model’s internal reasoning adapts as the input dimensionality increases from a radiometric baseline to a complex multi-modal fusion. Three primary experimental configurations are defined based on the hierarchical data stages:
\begin{itemize}[leftmargin=3em, itemsep=3pt,topsep=6pt]
    \item Baseline Stack ($C=13$): This setup establishes the core relationship between land cover and temperature, utilizing masked LST, one-hot encoded LULC (9 channels), and cyclical DOY encodings.
    \item Topographic-Vegetation Integration ($C=18$): This stage expands the baseline by introducing NDVI and primary terrain attributes (Elevation, Slope, Aspect) to provide context for vegetation-driven and altitude-dependent thermal lapse rates.
    \item Full Multi-Modal Fusion ($C=21$): This comprehensive feature set incorporates dynamic hillshade to account for solar-induced shading and dual-polarization Sentinel-1 SAR (VV/VH) to provide signals related to surface moisture and roughness.
\end{itemize}
Besides these input configurations, the framework is evaluated with a second generator in the SwinUNet-GAN variant, in which the FFC-based generator is replaced with a SwinUNet backbone~\cite{Cao.2021}, a U-shaped transformer with shifted-window self-attention, while the full multimodal input stack, the PatchGAN discriminator and the loss configuration remain unchanged.

\subsection{Implementation and Optimization}\label{ssec:implementation}
The high-dimensional complexity of the multimodal stack necessitated a departure from the high-speed fixed learning rates typically utilized in standard RGB inpainting. Preliminary testing demonstrated that the suggested baseline rates ($1 \times 10^{-3}$ for $G$ and $1 \times 10^{-4}$ for $D$) caused significant adversarial instability and mode collapse, largely due to the high variance introduced by the SAR and topographic input channels. To ensure stable convergence and preserve radiometric integrity, the following optimization protocol was implemented:
\begin{itemize}[leftmargin=3em, itemsep=3pt, topsep=6pt]
     \item Initial Rate Reduction: The initial learning rates were reduced to $2 \times 10^{-4}$ for the generator ($G$) and $5 \times 10^{-5}$ for the discriminator ($D$). This adjustment provided more stable optimization and prevented the discriminator from overpowering the generator during the early phases of multimodal feature fusion.

     \item Learning Rate Scheduling: To facilitate refined radiometric convergence, we utilized a \sloppy \texttt{ReduceLROnPlateau} scheduler with a decay factor of 0.5 and a patience of 5 epochs. This dynamic approach provided the necessary precision to minimize the $L_1$ reconstruction error once the global thermal structure was established, ensuring that the final output maintained high absolute temperature accuracy. We additionally evaluated cosine warmup scheduling, which provided smoother early-stage optimization and stable convergence behavior. However, experimental results on our dataset demonstrated that the \sloppy \texttt{ReduceLROnPlateau} scheduler achieved superior final performance for this specific training configuration.
\end{itemize}

The complete training configuration is summarized in Table~\ref{tab:training}. All models, including the configurations of the ablation study and the generator variants, were trained for 60 epochs on a single NVIDIA RTX 6000 Ada GPU (48 GB), which took approximately 36 h per model with the full multimodal input stack ($\approx$35 min per epoch including validation). The hyperparameters of this configuration were obtained by tuning, starting from the configuration proposed for LaMa~\cite{Suvorov.2021}, namely its learning rates, its ResNet-50 perceptual backbone and the loss weights $\kappa=10$, $\alpha=30$, $\beta=100$ and $\gamma=0.001$. Since tuning these settings with the original LaMa loss function alone did not reach a sufficient radiometric accuracy, the loss function was extended with the $L_1$ term ($\lambda=1$) and the gradient term ($\delta=1$). One setting was then varied at a time while all others were held fixed, and a change was retained only if it improved the training stability or the validation RMSE. In this way, the learning rates were reduced, the ReduceLROnPlateau scheduler was preferred over a fixed learning rate and over cosine annealing with warm-up, and the VGG-19 backbone was adopted, whereas none of the tested loss weights ($\kappa=15$, $\alpha=1$, 10 and 40, $\beta=30$ and 60) improved on the LaMa values. Further details on the individual loss terms and the perceptual backbone are given in Section~\ref{ssec:Phase3}.

\begin{table}[!t]
\centering
\footnotesize
\caption{Training configuration of the FFC-GAN and the SwinUNet-GAN variant with the full multimodal input stack. The settings tested during development are described in Section~\ref{ssec:implementation}.}
\label{tab:training}
\begin{tabular}{p{2.5cm}p{5.6cm}}
\toprule
Setting & Value \\
\midrule
Training data & 196 Landsat 8/9 Collection 2 Level-2 Tier 1 scenes of WRS-2 193/026 (2013--2025) with at most 80\% cloud cover, tiled into $256 \times 256$ pixels (30 m), of which only fully clear tiles are kept, split 80\%/10\%/10\% into training (29,094 tiles), validation (3,640 tiles) and test tiles \\
Training masks & $\approx$68,900 real Landsat cloud and cloud-shadow mask tiles (10\%--80\% masked), one drawn at random per sample \\
Input channels & 20 data channels and the binary mask \\
Generator parameters & 19.4 M (FFC-GAN), 28.0 M (SwinUNet-GAN) \\
Discriminator & PatchGAN, $N=4$, $\mathrm{ndf}=64$, updated every second iteration \\
Optimizer & Adam, $\beta_1 = 0.5$, $\beta_2 = 0.999$ \\
Learning rate ($G$/$D$) & $2 \times 10^{-4}$ / $5 \times 10^{-5}$ \\
Scheduler & ReduceLROnPlateau on the validation RMSE, factor 0.5, patience 5 epochs \\
Batch size, epochs & 16, 60 epochs \\
Model selection & Lowest validation RMSE (epoch 58 for the FFC-GAN, epoch 52 for the SwinUNet-GAN) \\
Loss weights & $\kappa=10$, $\alpha=30$, $\beta=100$, $\gamma=0.001$, $\lambda=1$, $\delta=1$ \\
Perceptual backbone & VGG-19 \\
Hardware, training time & One NVIDIA RTX 6000 Ada GPU (48 GB), $\approx$36 h for 60 epochs per model \\
\bottomrule
\end{tabular}
\end{table}

\subsection{Testing and Validation Protocol}
The models are tested on a set of 19 clear-sky Landsat scenes (cloud cover $<1\%$) acquired between 2015 and 2024. These scenes represent a significant computational challenge, with full-scene dimensions typically exceeding $6000 \times 6000$ pixels. To evaluate performance across varying degrees of data loss, each of the 19 scenes is independently masked by three categories of artificial cloud interference. The masks are real Landsat QA cloud masks of other acquisitions of the same path and row, which are selected according to the cloud cover reported in the metadata of their source scene, resulting in a total evaluation pool of 57 unique test cases ($19 \text{ scenes} \times 3 \text{ mask categories}$):
\begin{itemize}[leftmargin=3em,itemsep=0pt]
\item Low: 30.0\%--35.0\% cloud cover of the source scene.
\item Medium: 35.0\%--50.0\% cloud cover of the source scene.
\item High: 50.0\%--70.0\% cloud cover of the source scene.
\end{itemize}
Since each mask comprises not only the clouds but also the cloud shadows, dilated cloud, cirrus and snow pixels flagged in the QA band, and since it is combined with the gaps already present in the ground truth, the actual fraction of missing pixels of a test case is considerably larger than the cloud cover of the source scene. Across the 57 test cases, it ranges from 46\% to 85\% of the scene (median 61.5\%), with 19 cases exceeding 70\% and 2 cases exceeding 80\%, and the reconstruction accuracy is therefore additionally analyzed as a function of this actual missing fraction.
Performance is quantified using RMSE. Furthermore, a quantile analysis is conducted across the LST range (0\%--25\%, 25\%--50\%, 50\%--75\%, and 75\%--100\%) to identify potential biases at extreme temperature values, such as urban heat island hotspots.

\begin{figure}[h]
\centering
\scriptsize
\centering
\includegraphics[width =\columnwidth]{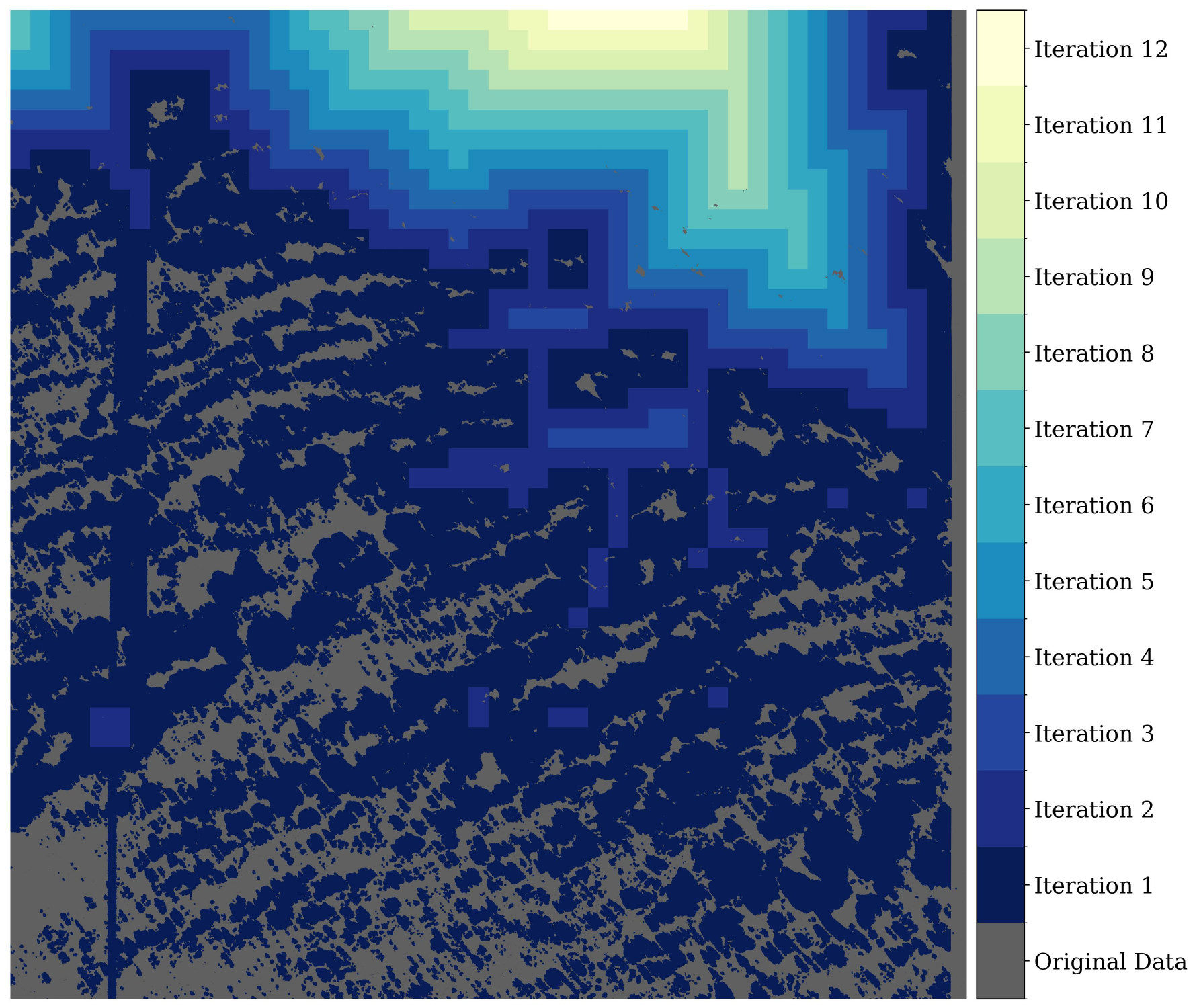}
\caption{Iterative inpainting progression for the masked regions on August 11, 2022.}
\label{fig:inpainting_sequence}
\end{figure}

For large-scale reconstruction of the full $6000 \times 6000$ pixel scenes, a sliding-window approach is employed using $256 \times 256$ tiles with a $128$-pixel stride ($50\%$ overlap). Standard sequential processing is prone to error accumulation in heavily masked regions. Therefore, to achieve the robust reconstructions under extreme data loss (>70\%) targeted in this study, an adaptive wave inpainting mechanism is implemented:

\begin{itemize}[leftmargin=3em,itemsep=3pt,topsep=6pt]
\item Sorting: Tiles are sorted based on their percentage of valid data.
\item Propagation:
The algorithm processes tiles with the most valid data first, using newly inpainted data as contextual information for overlapping neighbors. Fig.~\ref{fig:inpainting_sequence} visualizes this iterative process, demonstrating how the reconstruction spatially propagates inward from the valid data boundaries.
\item Quality Control: Tiles with less than $5\%$ valid data are skipped to prevent low-confidence estimations in heavily masked areas.
\end{itemize}

To evaluate the performance of the proposed adapted LaMa model for remotely sensed thermal data (specifically, LST), we compared it with two baselines and with the SwinUNet-GAN variant of our framework. First, the proposed approach is compared with the original LaMa~\cite{Suvorov.2021} model to isolate the impact of our thermal-specific adaptations. Second, the SwinUNet-GAN variant introduced in Section~\ref{ssec:variants} provides a rigorous architectural comparison to determine whether the global receptive field of Fast Fourier Convolutions handles complex thermal frequencies better than the hierarchical self-attention of a modern vision transformer. Finally, we include an XGBoost baseline inspired by the SDX-LST framework~\cite{Li.2024}. While the original SDX-LST successfully utilized SAR data to guide gap-filling beyond 70\% missing pixels, its application was limited to vegetated areas and tested on a relatively small scale. Adapting this XGBoost approach for a full Landsat scene provides a robust, traditional machine learning benchmark for our large-scale, heterogeneous landscape evaluation.

To assess the performance of the trained models outside the training region, the FFC-GAN and the SwinUNet-GAN variant are additionally applied, without any retraining or fine-tuning, to the independent region in western Germany (WRS-2 path 196, row 25) described in Section~\ref{ssec:Phase1}. For this region, 23 clear-sky Landsat 8/9 scenes acquired between 2015 and 2025 are used, of which 8 were acquired between October and March, compared with 5 of the 19 test scenes in the study area. All auxiliary inputs are prepared in the same way and normalized with the same parameters as for the study area. Artificial cloud masks of the same three categories are drawn from the same pool of cloud masks as in the study area and transferred onto the new scenes, which results in 69 test cases with actual missing fractions between 41\% and 89\% (median 59.7\%). In both regions, the two models are evaluated on identical masked test cases, so that their errors can be compared case by case.

\section{Results}\label{sec:Results}
\begin{figure}[!t]
\centering
\includegraphics[width=\columnwidth]{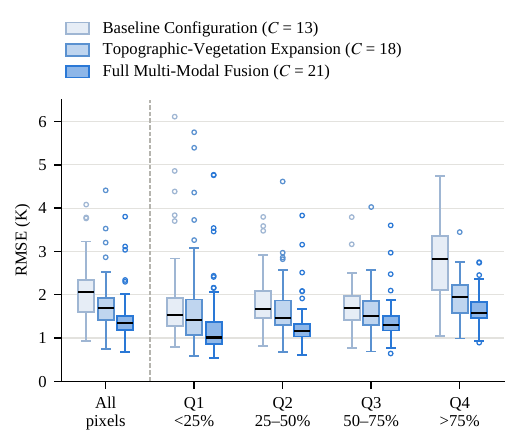}
\caption{Comparison of RMSE over all reconstructed pixels and across LST quantiles for successive guidance strategies in the Bavarian study area, evaluated on the same 57 masked test cases. Performance consistently improves with the inclusion of additional modalities, with the Full Multi-Modal Fusion demonstrating the lowest median errors and significantly reducing the large variance observed in the Q4 baseline. While the overall error distribution narrows, some isolated outliers remain observable representing challenging scene-level conditions and rare environmental cases within the dataset.}
\label{fig:ablation_study_results}
\end{figure}

\begin{figure*}[htbp]
\vspace*{-2cm}
\centering
\scriptsize
\centering
\includegraphics[width =\textwidth]{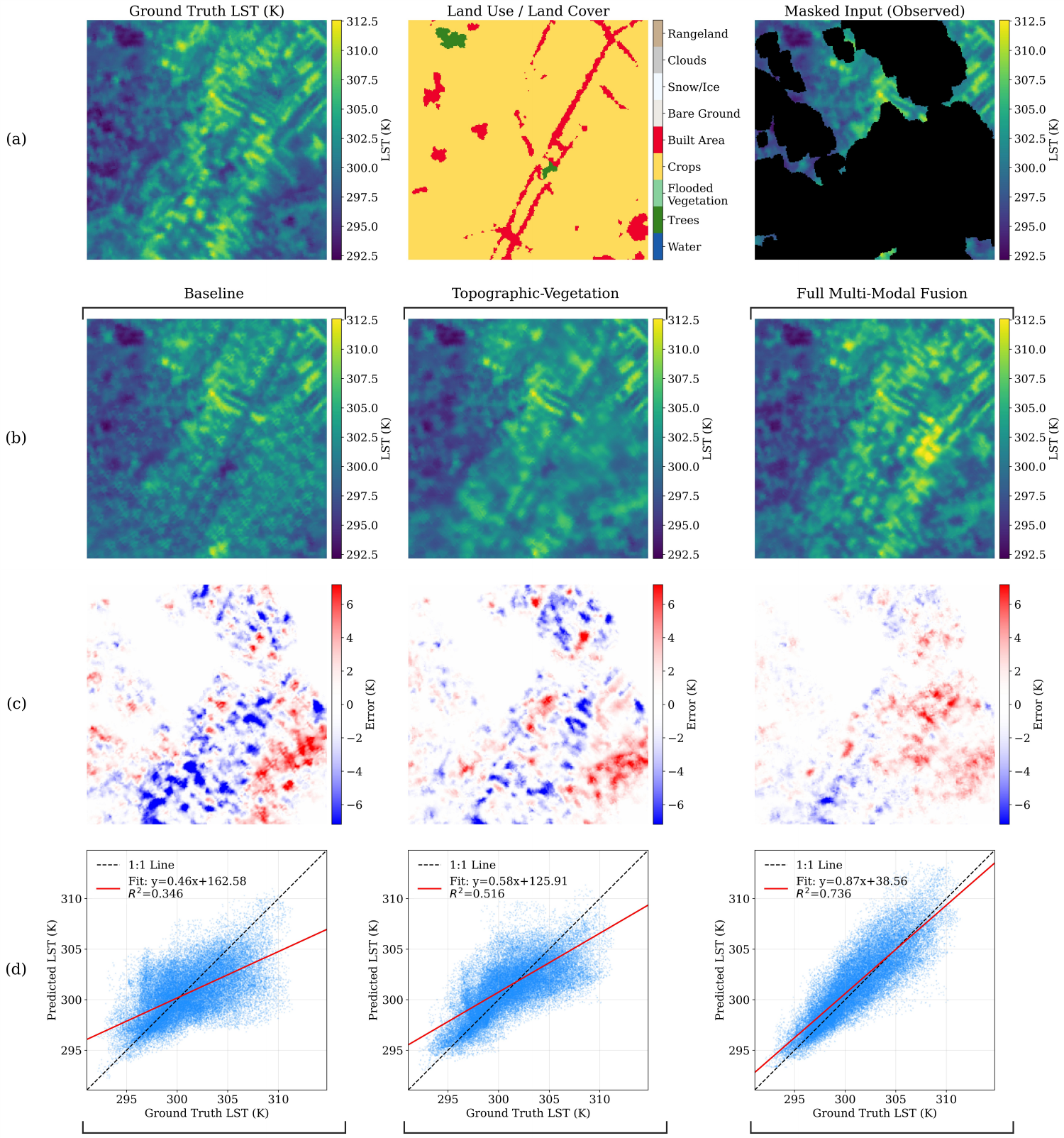}
\caption{Visual and statistical evaluation of LST reconstruction performance over a predominantly agricultural (crop) landscape. The figure illustrates the progressive improvement in spatial accuracy and correlation across the three model configurations: Baseline, Topographic-Vegetation, and Full Multi-Modal Fusion. (a) the top row displays the Ground Truth LST (K), the categorical Land Use / Land Cover map highlighting the spatial distribution of crops, and the Masked Input (Observed) showing the simulated regions of missing data in black. (b) Spatial LST (K) reconstructions generated by each respective model configuration. (c) Pixel-wise error maps representing the spatial distribution of the reconstruction error (Predicted LST minus Ground Truth LST in Kelvin). Red and blue clusters indicate areas of significant overestimation and underestimation, respectively, which are visibly suppressed in the Full Multi-Modal Fusion configuration. (d) Scatter plots showing the correlation between predicted and ground truth LST values for the reconstructed pixels. Each plot includes the 1:1 reference line (dashed black), the linear regression fit (solid red), and the coefficient of determination ($R^2$), demonstrating a substantial increase in predictive alignment (from $R^2=0.346$ to $R^2=0.736$) as multi-modal guidance is integrated.}
\label{fig:Performance_CropLand}
\end{figure*}

The results demonstrate a progressive reduction in reconstruction error as additional guidance modalities are incorporated. Fig.~\ref{fig:ablation_study_results} summarizes the statistical results (errors) of the three experimental configurations. The Baseline Configuration exhibits the highest RMSE values and the largest variability, particularly in the highest temperature quantile (Q4), where both the median error and spread increase substantially. A visual evaluation of the pixel-wise reconstruction error further revealed inaccuracies within vegetated regions. Fig.~\ref{fig:Performance_CropLand} demonstrates how the baseline configuration struggles with high-error clusters in crop areas, resulting in a weaker overall predictive alignment. Similar limitations are observed in heavily forested regions, detailed in Appendix~\ref{app:forested_terrain} (Fig.~\ref{fig:Performance_Trees}). These observations suggest that the LULC guidance alone was insufficient to capture the intra-class variability present in heterogeneous vegetation structures and varying surface conditions. To address this limitation, the Topographic-Vegetation Expansion strategy was introduced to provide additional environmental and structural context. This extension improves stability and reduces the overall RMSE distribution across most quantiles, although elevated errors remain present in Q4, and performance in vegetated areas still appears to have potential for further refinement. Ultimately, the Full Multi-Modal Fusion configuration achieves the best overall performance, producing the lowest median RMSE values and the tightest interquartile ranges across all quantiles. This improvement is especially pronounced in the lower and mid-temperature ranges (Q1--Q3), indicating enhanced reconstruction consistency. Although Q4 still exhibits comparatively higher errors due to the increased complexity of extreme thermal regions, the multimodal fusion approach significantly suppresses both variance and outlier magnitude relative to the other configurations. These results suggest that integrating complementary spatial and environmental guidance information substantially improves thermal reconstruction accuracy and robustness across varying LST conditions.

\begin{figure*}[!t] 
\vspace{1.1cm}
    \centering
    
    \begin{minipage}[t]{0.48\textwidth}
        \centering
        \subfloat[Ground Truth\label{fig:sub_a}]{
            \includegraphics[width=\textwidth]{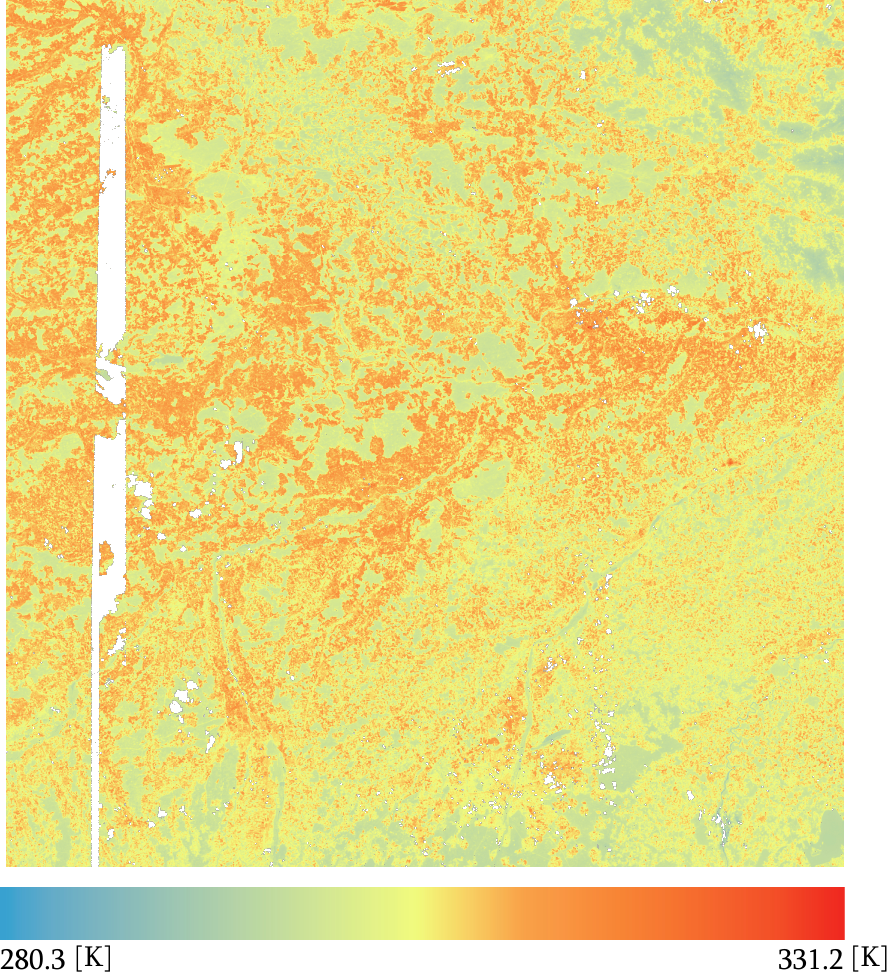}%
        }
    \end{minipage}%
    \hfill 
    \begin{minipage}[t]{0.48\textwidth}
        \centering
        \subfloat[Boolean Mask (Missing data = 1)\label{fig:sub_b}]{%
            \includegraphics[width=\textwidth]{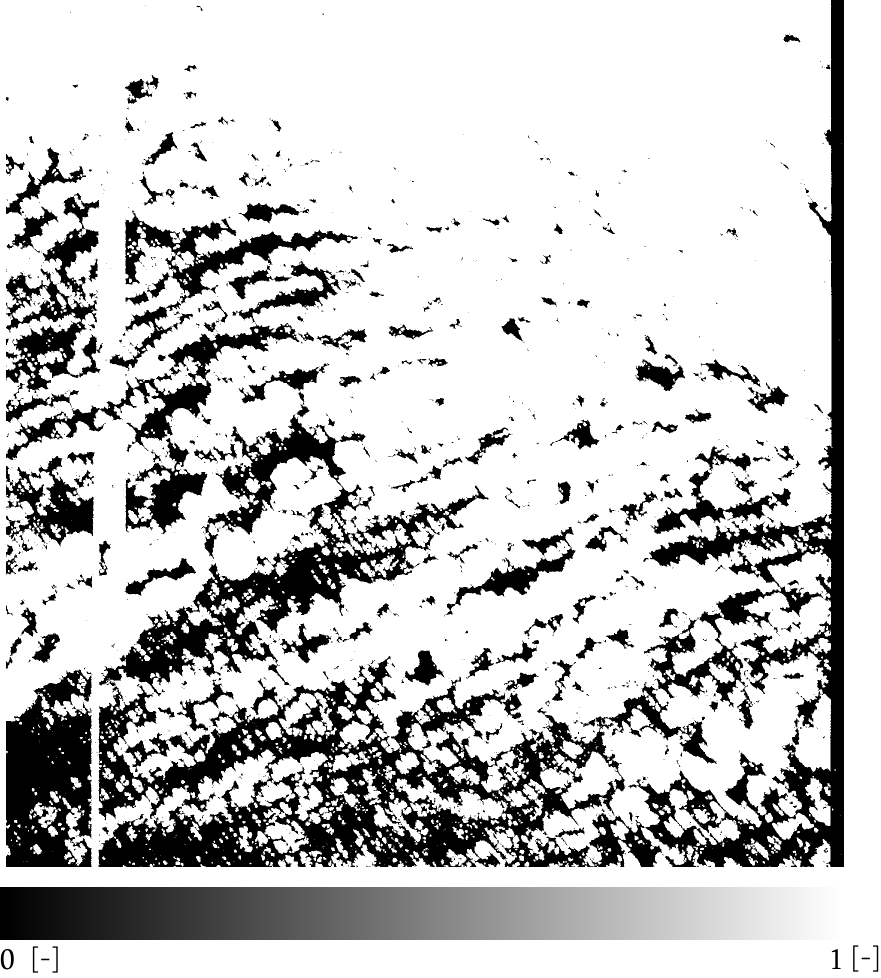}%
        }
    \end{minipage}
    
    \vspace{0.5em} 
    
    \begin{minipage}[t]{0.48\textwidth}
        \centering
        \subfloat[Estimated LST\label{fig:sub_c}]{%
            \includegraphics[width=\textwidth]{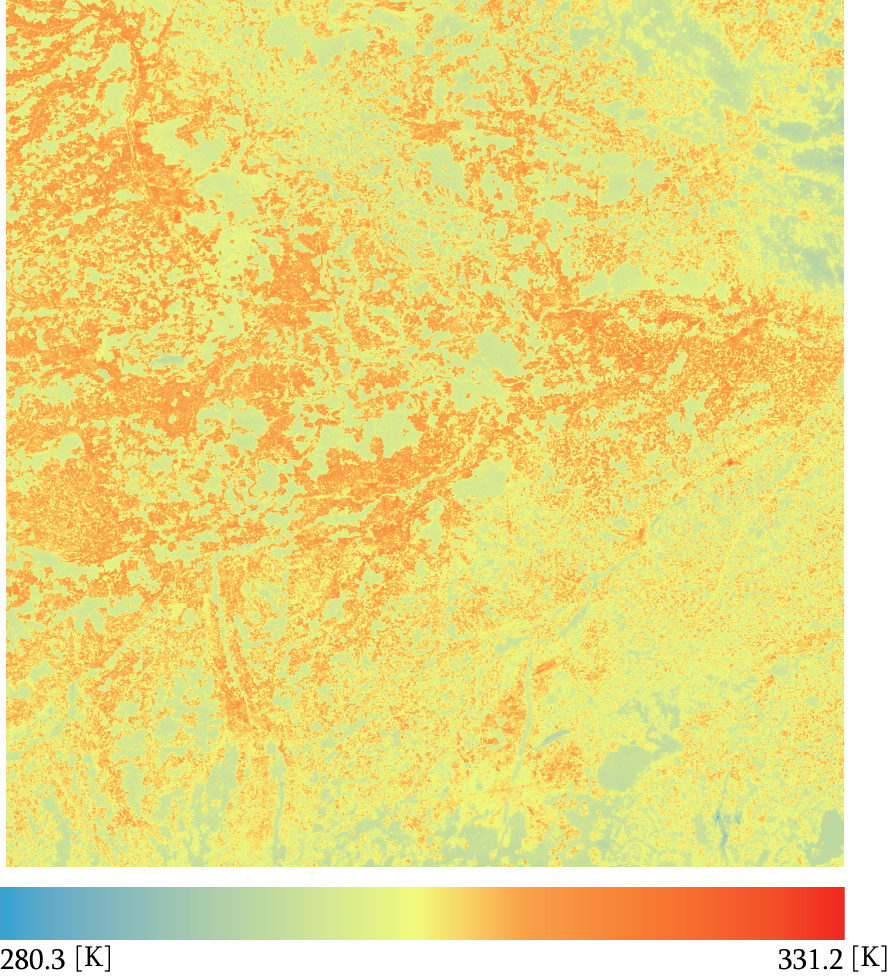}%
        }
    \end{minipage}%
    \hfill
    \begin{minipage}[t]{0.461\textwidth} 
        \centering
    
        \subfloat[Calculated error map\label{fig:sub_d}]{%
            \hspace*{-0.76cm}
            \includegraphics[width=\textwidth]{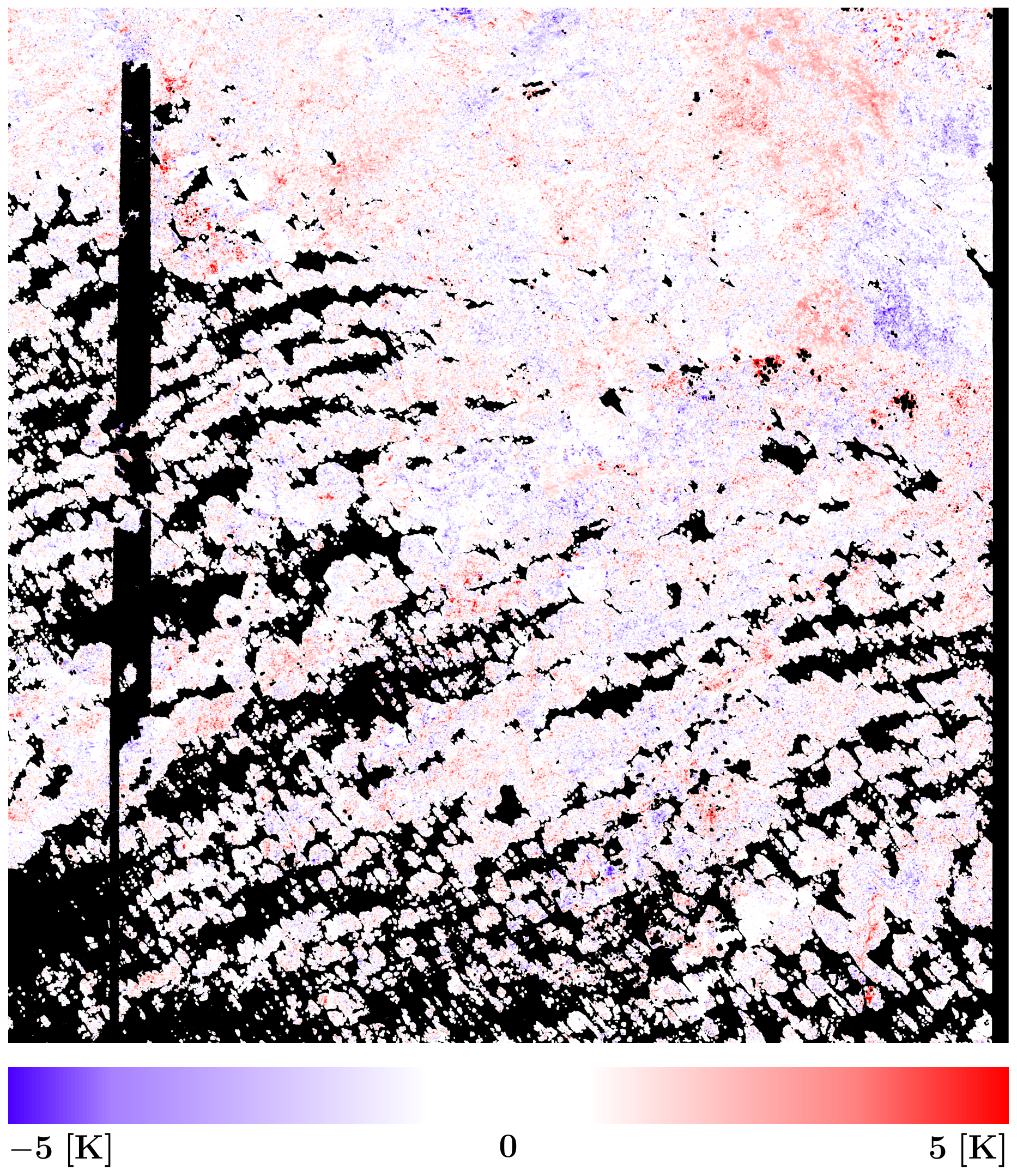}%
        }
        
        \vspace{0.1cm} 
        
        {\raggedright \footnotesize \setlength{\baselineskip}{1em} Black regions denote original cloud-free pixels or missing data lacking ground truth.\par}
    \end{minipage}
    
    \vspace{0.5cm} 
    
    \caption{Clear-Sky LST Reconstruction on August 11, 2022, using the proposed Full Multi-Modal architecture.}
    \label{fig:validation}
\end{figure*}

A representative example of the proposed architecture's performance in reconstructing an entire Landsat LST scene containing 77\% missing data is presented in Fig.~\ref{fig:validation}. The gaps in the ground truth image arise because Landsat LST products rely on the ASTER Global Emissivity Dataset (GED) provided by the Land Processes Distributed Active Archive Center (LP DAAC)~\cite{usgs_landsat_lst}. Consequently, any missing data in the ASTER GED inherently results in missing data within the Landsat LST product~\cite{usgs_landsat_lst}.

\begin{figure*}[!t]
    \centering
    \includegraphics[width=\textwidth]{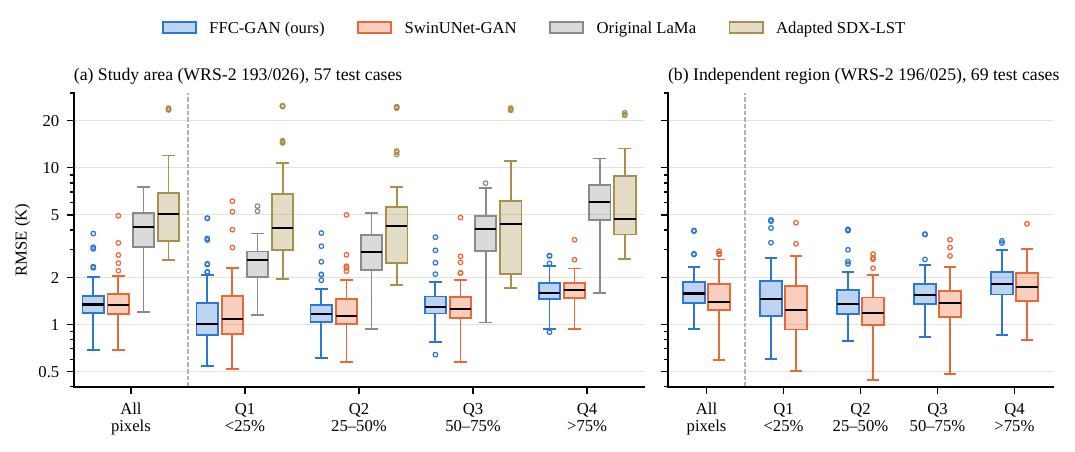}
    \caption{RMSE over all reconstructed pixels and across the four LST quantiles (a) of the original LaMa, the adapted SDX-LST, the SwinUNet-GAN variant and the FFC-GAN in the Bavarian study area (WRS-2 193/026) and (b) of the SwinUNet-GAN variant and the FFC-GAN in the independent region (WRS-2 196/025), to which both generator variants were applied without retraining. In each region, all models are evaluated on identical masked test cases. Note the logarithmic RMSE axis.}
    \label{fig:comparison}
\end{figure*}

To assess the proposed architecture and its optimal data stack, we benchmarked it against the original LaMa model alongside an adapted XGBoost framework derived from SDX-LST and the SwinUNet-GAN variant of our framework. The SDX-LST framework~\cite{Li.2024}, originally developed to reconstruct Landsat LST in vegetated regions using SAR and DEM auxiliary inputs, exhibits robust gap-filling capabilities even for gap sizes exceeding 70\%. For a fair comparison, its data stack was expanded to mirror our model's input configuration. Additionally, the Swin Transformer-enhanced GAN was selected due to the proven efficacy of its self-attention mechanisms in challenging remote sensing tasks, including cloud removal in 10-m satellite imagery~\cite{Zhu.2024} and image super-resolution~\cite{Huo.2025, Liu.2025}. We implemented this variant by replacing the FFC-based generator with a SwinUNet generator, maintaining the proposed full guiding data stack, discriminator architecture, and loss function. We quantified the performance of the three models using the same testing protocol applied to our proposed model. The only exception is the XGBoost-based framework, which does not require a sliding-window approach for large-scale reconstruction. The quantitative results of this comparison are presented in Fig.~\ref{fig:comparison}a. The comparison demonstrates that the adapted SDX-LST framework underperforms relative to the other models, with a median RMSE over all reconstructed pixels of 5.08~K, compared with 4.20~K for the original LaMa and 1.34~K and 1.33~K for the FFC-GAN and the SwinUNet-GAN. While the original LaMa model achieves a lower median RMSE than SDX-LST in the first three quantiles (Q1--Q3), its reconstruction accuracy drops significantly in the upper quartile (Q4), where its median RMSE (6.05~K) even exceeds that of SDX-LST (4.68~K). In contrast, the SwinUNet-GAN exhibits performance closely matching our proposed FFC-based model. However, it still produces larger outliers and exhibits wider error ranges.

\begin{figure*}[!t]
\centering
\includegraphics[width=\textwidth]{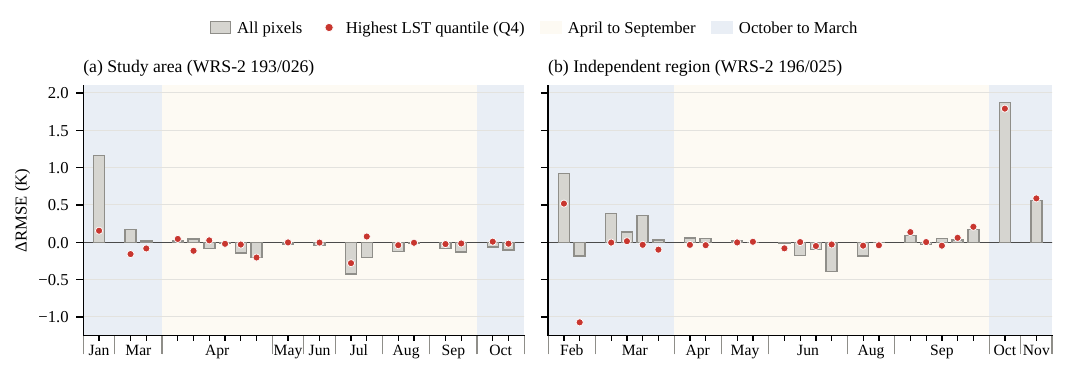}
\caption{Difference in RMSE between the FFC-GAN and the SwinUNet-GAN variant (FFC-GAN minus SwinUNet-GAN) per test scene (a) in the Bavarian study area and (b) in the independent region, averaged over the three mask categories of each scene, over all reconstructed pixels (bars) and in the highest LST quantile Q4 (markers). Negative values indicate a lower RMSE of the FFC-GAN. Each pin on the horizontal axis marks one scene, with the scenes acquired between 2015 and 2025 grouped by month irrespective of the year and ordered by day within each month. The background distinguishes acquisitions between April and September from those between October and March. Over all pixels, the FFC-GAN achieves a lower RMSE in 14 of the 19 scenes in the study area and in 9 of the 23 scenes in the independent region, and in Q4 it is more accurate in 39 of the 57 and in 38 of the 69 test cases.}
\label{fig:paired}
\end{figure*}

Fig.~\ref{fig:comparison}b shows the reconstruction errors of both generator variants in the independent region, which can be compared directly with those in the study area in Fig.~\ref{fig:comparison}a. Without retraining, the median RMSE of the FFC-GAN increases from 1.34~K in the study area to 1.58~K in the independent region, and that of the SwinUNet-GAN from 1.33~K to 1.39~K, while the median correlation coefficient of both variants decreases only slightly from 0.92 to 0.90. A scene-by-scene comparison on identical masks (Fig.~\ref{fig:paired}) shows that the FFC-GAN achieves a lower RMSE in most scenes of the study area, in particular in those acquired between April and September. In the independent region, both variants perform equally well between April and September, whereas the SwinUNet-GAN is more accurate in most scenes acquired between October and March, in which the FFC-GAN overestimates LST with a median bias of +0.68~K, compared with +0.02~K for the SwinUNet-GAN.

\begin{figure}[!t]
    \centering
    \includegraphics[width=\columnwidth]{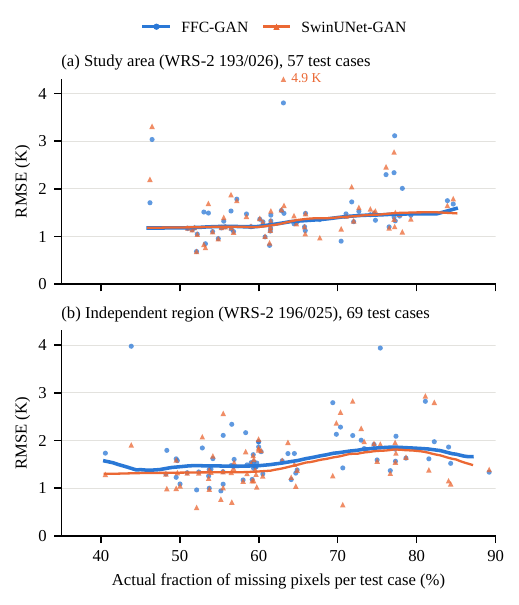}
    \caption{RMSE of the FFC-GAN and the SwinUNet-GAN variant as a function of the actual fraction of missing pixels per test case, which combines the artificial cloud mask with the gaps of the ground truth, (a) in the Bavarian study area (57 test cases) and (b) in the independent region (69 test cases). Markers show the individual test cases and lines the running median within a moving window of $\pm$10 percentage points. One test case of the SwinUNet-GAN in the study area (4.9~K) lies above the range of the axis and is drawn at its upper edge.}
    \label{fig:missing}
\end{figure}

Fig.~\ref{fig:missing} relates the RMSE of every test case to its actual fraction of missing pixels, which ranges from 46\% to 85\% in the Bavarian training area and from 41\% to 89\% in the independent region in western Germany. In both regions, the running median of the RMSE increases only moderately with the missing fraction and remains below 2~K over the entire range, although individual test cases exceed this value. When the three masks of the same scene are compared, an increase of the missing fraction by 10 percentage points raises the RMSE of the FFC-GAN by only 0.05~K in Bavaria and 0.11~K in the independent region, which indicates that the differences between the test cases are mainly caused by the conditions of the individual scenes rather than by the size of the gaps.

\begin{figure*}[!t]
    \centering
    \includegraphics[width=\textwidth]{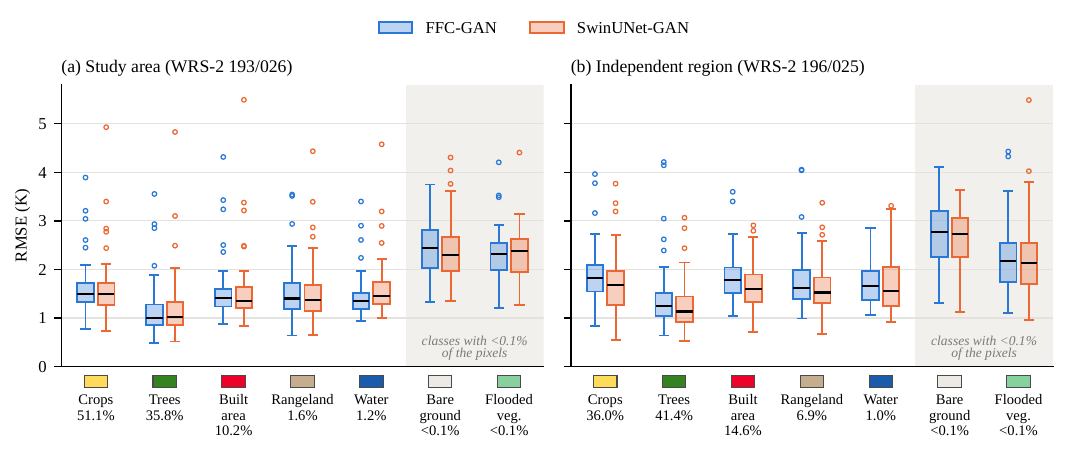}
    \caption{RMSE per LULC class of the FFC-GAN and the SwinUNet-GAN variant (a) in the Bavarian study area and (b) in the independent region. Each box summarizes the RMSE of all reconstructed pixels of a class per test case, and the percentages give the share of each class among all reconstructed pixels. The class clouds of the LULC map, which covers less than 0.01\% of the pixels, is not shown. The colors of the classes correspond to the land-cover maps in Figs.~\ref{fig:Performance_CropLand} and~\ref{fig:urban}, and the shaded area marks the classes with less than 0.1\% of the reconstructed pixels.}
    \label{fig:lulc}
\end{figure*}

To assess the reconstruction accuracy across surface types, the errors of all reconstructed pixels are further grouped by the classes of the LULC map used as model input (Fig.~\ref{fig:lulc}). In both regions and for both generator variants, trees exhibit the lowest errors, with median RMSE values between 1.0~K and 1.25~K, followed by crops, built area, rangeland and water with median values between 1.35~K and 1.8~K, whereas bare ground and flooded vegetation show the highest errors of 2.1~K to 2.8~K but together account for less than 0.1\% of the reconstructed pixels. In Bavaria, the FFC-GAN achieves a lower RMSE than the SwinUNet-GAN in the majority of test cases for crops, trees, built area and water, whereas in western Germany the SwinUNet-GAN is more accurate in the majority of test cases for crops, trees, built area and rangeland, which is consistent with the overall comparison of both variants.

\begin{figure*}[!t]
    \centering
    \includegraphics[width=0.97\textwidth]{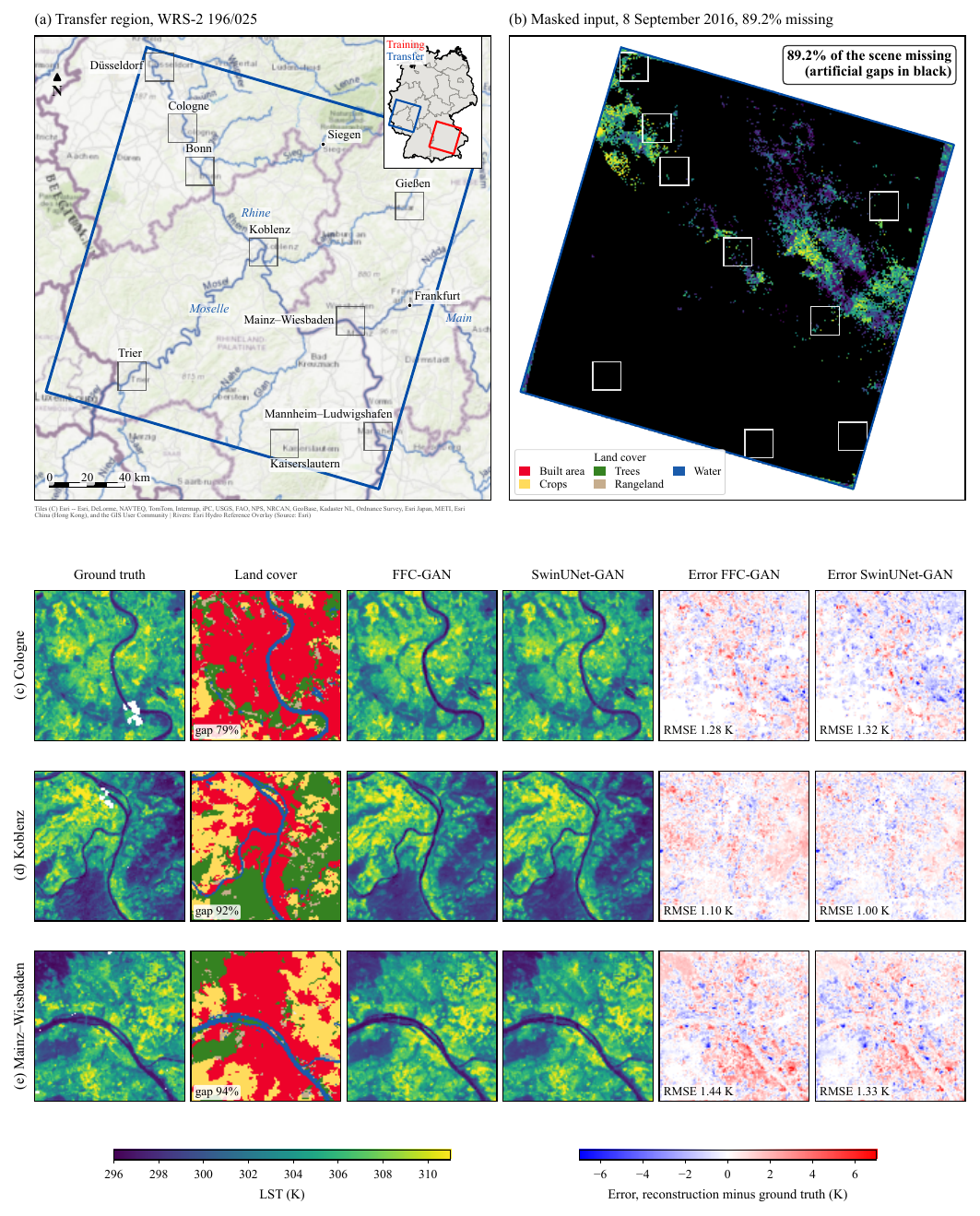}
    \caption{Urban reconstruction examples in the independent region (WRS-2 196/025) for the scene of September 8, 2016 with 89.2\% missing data. (a) Evaluated scene (blue outline), city windows and location of the training and transfer regions (inset), basemap: Esri World Topographic Map with the Esri Hydro Reference Overlay. (b) Masked input with the artificial gaps in black. (c)--(e) Ground truth, land cover, reconstructions of the FFC-GAN and the SwinUNet-GAN, and their errors (reconstruction minus ground truth) in 15~km~$\times$~15~km windows over Cologne, Koblenz and Mainz--Wiesbaden. The share of each window within the gaps and the RMSE over the reconstructed pixels are given in the panels.}
    \label{fig:urban}
\end{figure*}

Since urban areas are of particular interest because sealed and bright surfaces generate high LST, Fig.~\ref{fig:urban} presents reconstruction examples for cities of the independent region in the scene of September 8, 2016, in which 89.2\% of the scene is missing. The 15~km~$\times$~15~km windows are placed on cities whose ground truth is complete and which lie largely or entirely within the artificial gaps, and six further cities of the same scene are shown in Fig.~\ref{fig:urban2} in Appendix~\ref{app:urban}. Both generator variants reproduce the warmer built-up areas as well as the cooler corridors of the Rhine and Moselle rivers, and the RMSE within the nine city windows ranges from 1.10~K to 1.97~K for the FFC-GAN and from 1.00~K to 1.87~K for the SwinUNet-GAN, which is comparable to the RMSE of the entire scene (1.33~K and 1.39~K).

\section{Discussion}\label{sec:Discussion}

The accurate reconstruction of 30~m LST under persistent and extensive cloud cover remains a critical bottleneck in thermal remote sensing. This study demonstrates that transitioning from local convolutional architectures to an FFC-based network, guided by a high-dimensional physical feature stack, successfully recovers both structural and radiometric thermal patterns even when over 70\% of a Landsat scene is obscured. By avoiding the dependency on gap-filled optical proxies at the Landsat scale and on temporally adjacent image pairs, the proposed framework establishes a new pathway for large-scale thermal inpainting in highly heterogeneous environments. 

\subsection{Contribution of the Auxiliary Inputs}
The evaluation demonstrates the progressive reduction in reconstruction error achieved by moving beyond standard land-cover constraints. The ablation study, summarized in Fig.~\ref{fig:ablation_study_results}, illustrates that relying solely on LULC and cyclical time encodings (Baseline Configuration) is insufficient to capture the high intra-class variance of LST, particularly in complex urban hotspots and heterogeneous vegetation. The substantial performance improvement observed in the Full Multi-Modal Fusion configuration demonstrates the value of integrating physically grounded environmental variables: dynamic hillshade and SAR data. The inclusion of dynamic hillshade allowed the network to differentiate between static altitude-dependent lapse rates and transient, solar-induced shading patterns. Furthermore, the integration of cloud-penetrating SAR data allows neural networks to utilize microwave backscatter variations to infer complex physical variables, including vegetation structure, soil moisture, and surface roughness~\cite{Tsokas.2022}. Among these, soil moisture is a key physical variable governing the thermal inertia of the environment~\cite{Cheruy.2017}, thereby providing the model with a physically grounded representation of landscape thermodynamics. Overall, this multi-modal integration grounds the generator’s spectral reasoning in the underlying geophysical state of the landscape. These findings suggest that explicitly incorporating geophysical priors is critical for modeling LST variability, and help explain why a remote-sensing-adapted version of LaMa is better suited to this task than the original formulation, which is not designed to exploit physically grounded environmental signals.


\subsection{Generator Variants and Performance in the Independent Region}
Further insights are obtained through benchmarking against the SwinUNet-GAN variant of our approach and an XGBoost model adapted from the established SDX-LST framework~\cite{Li.2024}. While the adapted SDX-LST framework utilizes the same auxiliary inputs, its tree-based approach struggled to extrapolate thermal gradients across the massive spatial gaps (>70\%) typical of full Landsat scenes. Furthermore, while the SwinUNet-GAN provided highly competitive performance, the FFC-based generator proved superior in stabilizing outlier predictions, particularly in the extreme upper thermal quartiles (Q4), in Bavaria, where the training data were also obtained. This suggests that, under the conditions represented in the training data, the image-wide receptive fields enabled by spectral transformations (FFCs) are more adept at mapping distal topographic and environmental correlations than the hierarchical self-attention of the Swin Transformer. In the independent region in western Germany, however, the SwinUNet-GAN transferred more robustly, and the scene-by-scene comparison shows that this difference arises almost entirely from the cold-season scenes, in which the FFC-GAN systematically overestimates LST, while both variants perform equally well in the warm season. Since the training pool only admits fully clear-sky tiles, and such conditions occur predominantly in the warm season in Central Europe, cold-season conditions are underrepresented in the training data, which is consistent with the warm bias of the FFC-GAN in these scenes. The choice of the generator therefore represents a trade-off within the same framework, in which the FFC-GAN achieves the highest accuracy in Bavaria, particularly for the hottest pixels, with about 30\% fewer parameters (19.4~M compared with 28.0~M), whereas the SwinUNet-GAN generalizes more robustly to cold-season conditions in an unseen region. Since both generators share the multimodal input stack, the loss configuration and the full-scene inpainting strategy, these results also indicate that the proposed framework is not tied to a specific backbone, and season-balanced training or regional fine-tuning represent promising directions to further reduce the cold-season bias.

\subsection{Data Dependencies and Applicability}
In terms of spatial generalization, the framework is designed to be retrained easily for new regions rather than to be applied directly to any region, and its application without retraining to an independent region with a comparable climate increased the median RMSE only moderately from 1.34~K to 1.58~K for the FFC-GAN and from 1.33~K to 1.39~K for the SwinUNet-GAN. A primary constraint of many existing high-resolution LST reconstruction models is their reliance on clear-sky optical proxies at the Landsat scale, such as Landsat NDVI, which themselves have to be gap-filled before they can guide the reconstruction~\cite{Zhu.2022}, and Landsat NDVI time series require dedicated reconstruction methods before they can serve as predictors~\cite{Chen.2021}. Similarly, fusion-based approaches require the gap-filling of their reference images~\cite{Luo.2020, Tang.2023, Zhao.2026a}, the intercalibration of coarse and fine sensors~\cite{Yang.2022, Chen.2023} or the deep-learning downscaling of coarse LST~\cite{Daniels.2023}. Our framework also relies on optical information, namely NDVI and land cover, but takes both from globally available operational products, the 300 m NDVI composite of the Copernicus Land Monitoring Service and the 10 m LULC map, which are only resampled to the 30 m Landsat grid, while the remaining inputs consist of microwave SAR backscatter, digital elevation models and derived topographic indices. The framework therefore does not remove the dependency on optical data, but it avoids the reconstruction of optical proxies at the Landsat scale before the LST reconstruction. Furthermore, the selected Bavarian study area provides an excellent testbed due to its landscape heterogeneity, encompassing dense urban cores, industrial hubs, agricultural plains, and complex alpine topography.
Because all inputs of the framework can be downloaded as ready-to-use products that only need to be resampled to the Landsat grid, and since the code is openly available, the model can be retrained or fine-tuned for other regions with little preprocessing effort.

\subsection{Limitations}
It is important to acknowledge the limitations of the validation approach with respect to the broader objectives of thermal reconstruction. The experimental design utilizes synthetic cloud masks applied to clear-sky imagery, a standard methodology that allows for rigorous, pixel-to-pixel quantitative validation against true observations. Consequently, the proposed FFC-GAN model is specifically optimized to reconstruct the clear-sky LST which represents the theoretical baseline thermal state of the landscape in the absence of atmospheric obscuration. This approach inherently differs from estimating all-weather LST, where the physical presence of clouds alters the surface energy balance by reducing incoming shortwave radiation and trapping longwave emissions, thereby lowering the actual under-cloud kinetic temperature relative to clear-sky conditions during the day. Nevertheless, the clear-sky LST serves its own applications, and the framework can be extended to all-weather LST, either as the clear-sky step of a two-step framework or by adding fields that contain the cloud effect to its guiding stack. However, the accuracy of such an extension at 30~m cannot yet be assessed reliably, since no reference beyond sparse point measurements exists under clouds, as outlined in Section~\ref{sec:Introduction}. Users should keep in mind that clear-sky values exceed all-weather values by about 2~K to 8~K during the day at mid-latitudes~\cite{Ermida.2019} and that interpreting them as all-sky conditions can lead to an overestimation of surface urban heat islands~\cite{Yang.2024}.

\section{Conclusion}\label{sec:Summary}
This paper presents a multi-modal adaptation of the LaMa architecture for high-resolution LST reconstruction under extensive cloud cover. The proposed framework recovers spatially consistent thermal fields at 30 m resolution from partially observed Landsat scenes, establishing a new pathway for filling large gaps in fine-spatial-resolution remote sensing imagery. 
The framework leverages auxiliary inputs derived from readily available public Earth observation datasets and is entirely built in Python using standard geospatial and deep learning libraries. The code is released as open-source, enabling reproducibility and application to other remote sensing datasets and geographic regions, as the model can be re-trained using publicly available datasets.

The choice of the LaMa architecture, a GAN-based inpainting framework with an FFC-enhanced generator, addresses limitations of conventional deep learning models that rely on locally constrained receptive fields. Furthermore, the systematic integration of auxiliary inputs introduces physically informed guidance to the reconstruction process while maintaining practical accessibility and scalability across diverse regions. In addition, the LaMa framework is adapted for remote sensing data through targeted modifications to the training pipeline, including adjustments to the loss function design, learning rate strategy, and discriminator update schedule, enabling more effective learning given the statistical properties and spatial heterogeneity of LST fields.

To demonstrate the effectiveness of the proposed approach, the framework was systematically evaluated through ablation studies and benchmarking against established baselines. The findings show consistent improvements in reconstruction accuracy when incorporating SAR and topographic information, particularly under extreme spatial sparsity conditions corresponding to more than 70\% missing data in full Landsat scenes ($\sim6000\times6000$ pixels). 
This setting reflects large-scale, scene-level inpainting, where long-range spatial dependencies and global structural consistency are critical. Furthermore, the evaluation demonstrates improved stability in high-temperature ranges and reduced sensitivity to outlier predictions in the upper LST distribution. In addition, the application to an independent region without retraining resulted in only a moderate increase in reconstruction error, and the comparison with the SwinUNet-GAN variant showed that the framework can be combined with different generator backbones, where the FFC-based generator is most accurate in the study area and the transformer-based generator transfers more robustly to cold-season conditions. Overall, the results demonstrate that combining frequency-domain global receptive fields with physically grounded environmental predictors improves the reconstruction of spatially complex thermal patterns and enables more consistent representation of surface temperature variability across heterogeneous landscapes.

The proposed framework is the first, to the best of our knowledge, to combine FFC-based generative modeling with SAR-derived physical constraints for large-scale LST reconstruction in a unified learning environment, enabling consistent thermal inpainting over highly heterogeneous and partially observed satellite scenes.

\section*{Data Availability}
\label{sec:DataAvailability}
Remote sensing data used in this study are publicly available. The data sources are cited within the text, and full dataset details are provided in the reference list. 

\section*{Code Availability}
The source code for data preparation, training and full-scene inference, the configuration files and the trained generators of the FFC-GAN and the SwinUNet-GAN are available at \url{https://github.com/Marwa-AL/Landsat-LST-Reconstruction}. The repository also documents how to download and prepare all auxiliary datasets, and the prepared training tiles of the study area are available at \url{https://huggingface.co/datasets/Marwa-AL/FFC-GAN-LST-Bavaria}.

\section*{Conflict of interest}
The authors declare no conflicts of interest.

\appendices
\section{Visual Evaluation of Thermal Reconstruction in Forested Terrain}
\label{app:forested_terrain}
\setcounter{figure}{0}
\renewcommand{\thefigure}{A.\arabic{figure}}
To further validate the improvements observed in agricultural regions, a supplementary visual evaluation was conducted over a forested landscape. Fig.~\ref{fig:Performance_Trees} presents the resulting LST reconstructions and their corresponding spatial error distributions. Consistent with observations from croplands, relying solely on LULC guidance in the Baseline Configuration proves insufficient for accurately modeling structurally complex forested ecosystems. The categorical nature of LULC fails to account for thermal heterogeneity. Incorporating the Topographic-Vegetation Expansion begins to resolve some of this intra-class variability. This intermediate configuration provides better environmental context, which reduces the magnitude of the error clusters and improves the general alignment of predictions along the 1:1 reference line (Fig.~\ref{fig:Performance_Trees}(d)). However, the greatest improvements are realized under the Full Multi-Modal Fusion configuration. By integrating structural proxies such as SAR and hillshade, the model captures the fine-scale thermal gradients driven by the forested terrain's physical structure. The resulting error maps (Fig.~\ref{fig:Performance_Trees}(c)) demonstrate a minimization of localized inaccuracies, yielding a more uniform spatial error distribution. Furthermore, the corresponding scatter plot (Fig.~\ref{fig:Performance_Trees}(d)) for the full fusion approach exhibits the tightest data grouping and the strongest linear relationship among all configurations tested. These visual results reinforce the conclusion that integrating complementary multi-modal guidance is critical for robust and accurate thermal reconstruction in heterogeneous vegetated environments.

\begin{figure*}[p]
    \vspace*{-1.5cm} 
    \centering
    \includegraphics[width=\textwidth]{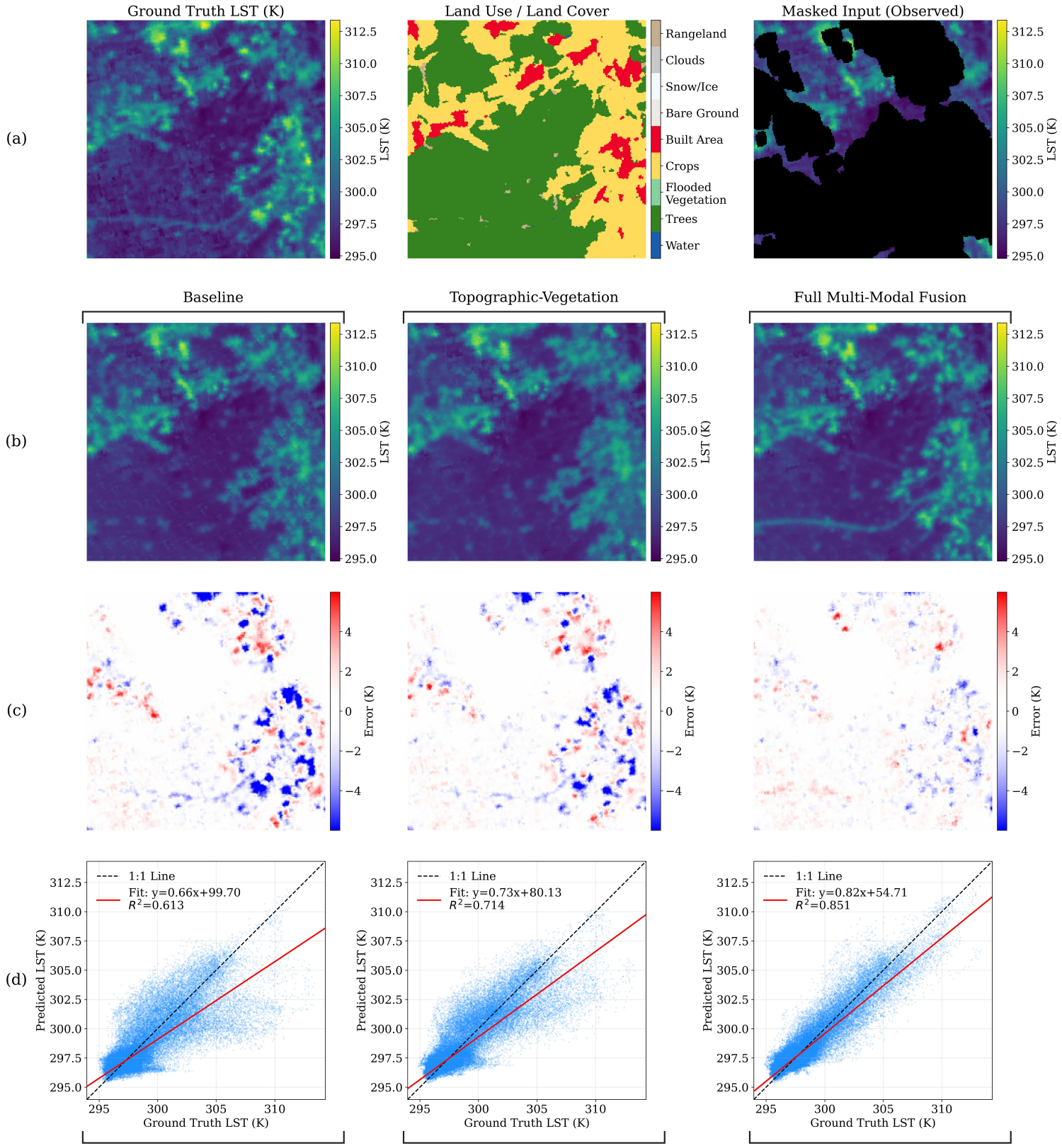}
    \caption{\footnotesize Visual and statistical evaluation of Land Surface Temperature (LST) reconstruction performance over a heavily forested (tree-dominated) landscape, illustrating the progressive improvement in spatial accuracy and correlation across three model configurations. \textbf{(a)} The top row displays the Ground Truth LST (K), the categorical Land Use / Land Cover map highlighting the spatial distribution of trees, and the Masked Input (Observed) showing the simulated regions of missing data in black. \textbf{(b)} Spatial LST (K) reconstructions generated by the Baseline, Topographic-Vegetation, and Full Multi-Modal Fusion configurations. \textbf{(c)} Pixel-wise error maps representing the spatial distribution of the reconstruction error (Predicted LST minus Ground Truth LST in Kelvin). Substantial spatial error clusters (red and blue) present in the Baseline model are visibly suppressed in the Full Multi-Modal Fusion configuration. \textbf{(d)} Scatter plots displaying the correlation between predicted and ground truth LST values. Each plot includes the 1:1 reference line (dashed black), the linear regression fit (solid red), and the coefficient of determination, demonstrating a significant increase in predictive alignment (from $R^2=0.613$ to $R^2=0.851$) as multi-modal guidance is integrated.}
    \label{fig:Performance_Trees}
\end{figure*}

\section{Further Urban Reconstruction Examples}
\label{app:urban}
\setcounter{figure}{0}
\renewcommand{\thefigure}{B.\arabic{figure}}
Fig.~\ref{fig:urban2} shows six further cities of the scene of September 8, 2016 in the independent region, selected with the same criteria and presented in the same layout as Fig.~\ref{fig:urban}.

\begin{figure*}[p]
    \centering
    \includegraphics[width=\textwidth]{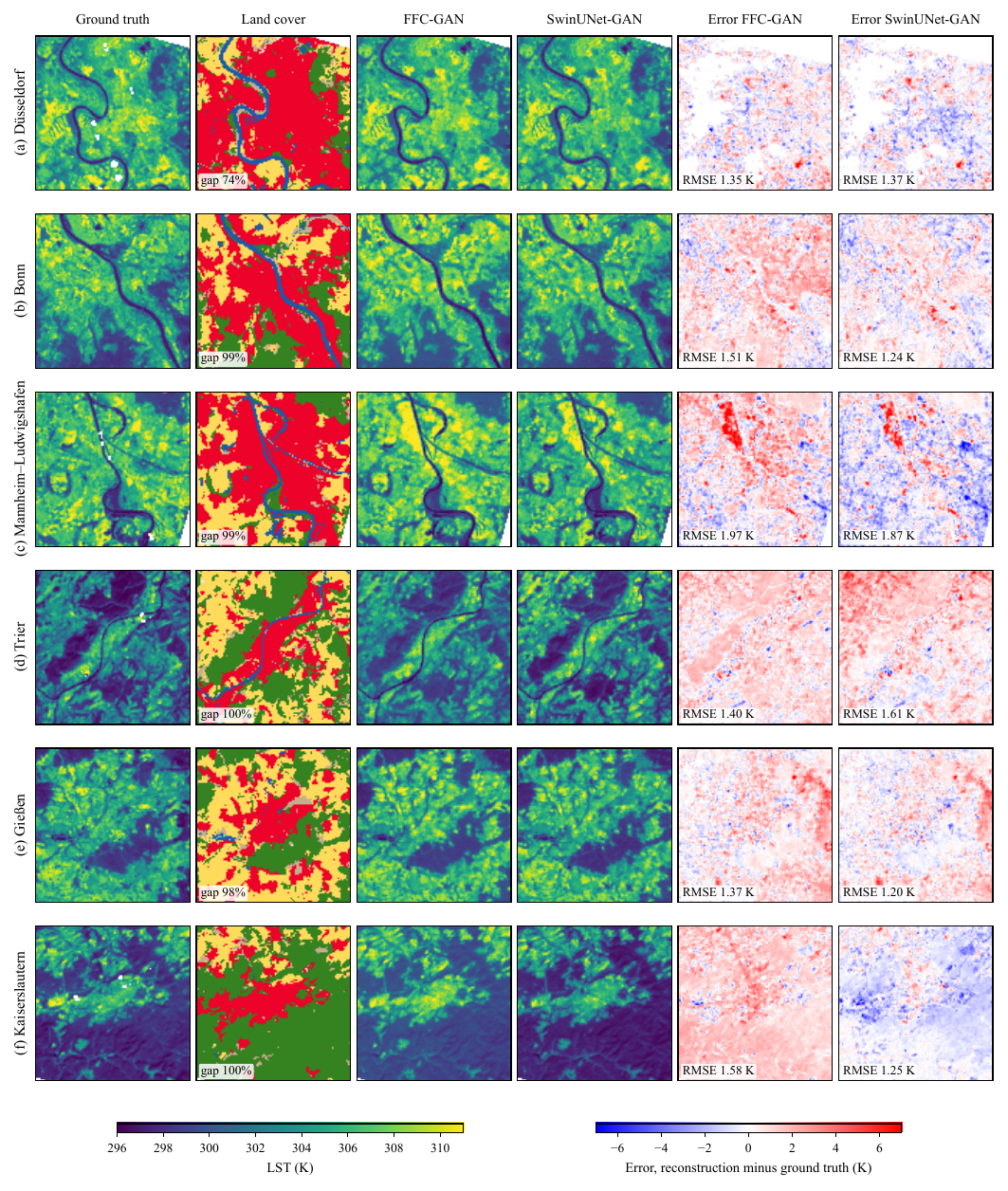}
    \caption{Further urban reconstruction examples in the independent region for the scene of September 8, 2016 with 89.2\% missing data: (a) Düsseldorf, (b) Bonn, (c) Mannheim--Ludwigshafen, (d) Trier, (e) Gießen and (f) Kaiserslautern. Columns and color scales as in Fig.~\ref{fig:urban}.}
    \label{fig:urban2}
\end{figure*}

\section*{Acknowledgment}
\label{sec:Acknowledgment}
During the preparation of this work, the authors used Claude (Anthropic) to improve the English language and readability of the manuscript and to assist in generating figures. All AI-assisted text and figures were reviewed, edited, and verified by the authors, who take full responsibility for the content of this publication. 

\clearpage
\bibliographystyle{IEEEtran}
\bibliography{bibliography_revised}

\vfill

\end{document}